\documentclass[10pt,journal,compsoc]{IEEEtran}

\ifCLASSOPTIONcompsoc
  \usepackage[nocompress]{cite}
\else
  \usepackage{cite}
\fi

\ifCLASSINFOpdf
  \usepackage[pdftex]{graphicx}
  \DeclareGraphicsExtensions{.pdf,.jpg,.jpeg,.png}
\else
  \usepackage[dvips]{graphicx}
  DeclareGraphicsExtensions{.eps}
\fi

\usepackage{amsmath}

\usepackage{algorithm}
\usepackage[noend]{algpseudocode}
\usepackage{multirow}

\usepackage[T1]{fontenc}
\usepackage{subcaption}
\usepackage[numbers]{natbib}
\usepackage{changes}
\usepackage{todonotes}
\usepackage{xspace}
\usepackage{booktabs}
\usepackage[colorlinks=true, allcolors=black]{hyperref}

\newcommand\yao[1]{\textcolor{black}{#1}}

\newcommand{\methodSaliencyNameLong}{Multi-Duration Element Attention Model~(MD-EAM)\xspace}
\newcommand{\methodSaliencyNameShort}{MD-EAM\xspace}

\newcommand{\methodNameLong}{Unified Model of Saliency and Scanpaths (UMSS)}
\newcommand{\methodNameShort}{UMSS\xspace}

\begin{document}

\title{Scanpath Prediction on Information Visualisations}

\author{Yao~Wang,~Mihai~B\^{a}ce,~and~Andreas~Bulling%
\IEEEcompsocitemizethanks{\IEEEcompsocthanksitem
Yao~Wang,~Mihai~B\^{a}ce,~and~Andreas~Bulling are with the Institute for Visualisation and Interactive Systems, University of Stuttgart, Germany. \protect\\
E-mail \{yao.wang, mihai.bace, andreas.bulling\}@vis.uni-stuttgart.de.
\IEEEcompsocthanksitem Yao Wang is the corresponding author.}%
\thanks{Manuscript received xx xx, 2022; revised xx xx, 2022.}
}

\markboth{IEEE Transactions on Vizualisation and Computer Graphics}%
{Wang \MakeLowercase{\textit{et al.}}: Scanpath Prediction on Information Visualisations}

\IEEEtitleabstractindextext{%
\begin{abstract}
    We propose \methodNameLong -- a model that learns to predict multi-duration saliency and scanpaths~(i.e. sequences of eye fixations) on information visualisations.
Although scanpaths provide rich information about the importance of different visualisation elements during the visual exploration process, prior work has been limited to predicting aggregated attention statistics, such as visual saliency.
We present in-depth analyses of gaze behaviour for different information visualisation elements~(e.g. Title, Label, Data) on the popular MASSVIS dataset.
We show that while, overall, gaze patterns are surprisingly consistent across visualisations and viewers, there are also structural differences in gaze dynamics for different elements.
Informed by our analyses, \methodNameShort~first predicts multi-duration element-level saliency maps, then probabilistically samples scanpaths from them.
Extensive experiments on MASSVIS show that our method consistently outperforms state-of-the-art methods with respect to several, widely used scanpath and saliency evaluation metrics.
Our method achieves a relative improvement in sequence score of 11.5\,\% for scanpath prediction, and a relative improvement in Pearson correlation coefficient of up to 23.6\,\% for saliency prediction.
These results are auspicious and point towards richer user models and simulations of visual attention on visualisations without the need for any eye tracking equipment.

\end{abstract}

}

\maketitle

\IEEEdisplaynontitleabstractindextext

\IEEEpeerreviewmaketitle

\IEEEraisesectionheading{\section{Introduction}}
\begin{figure*}[t]
    \centering
    \includegraphics[width=\textwidth]{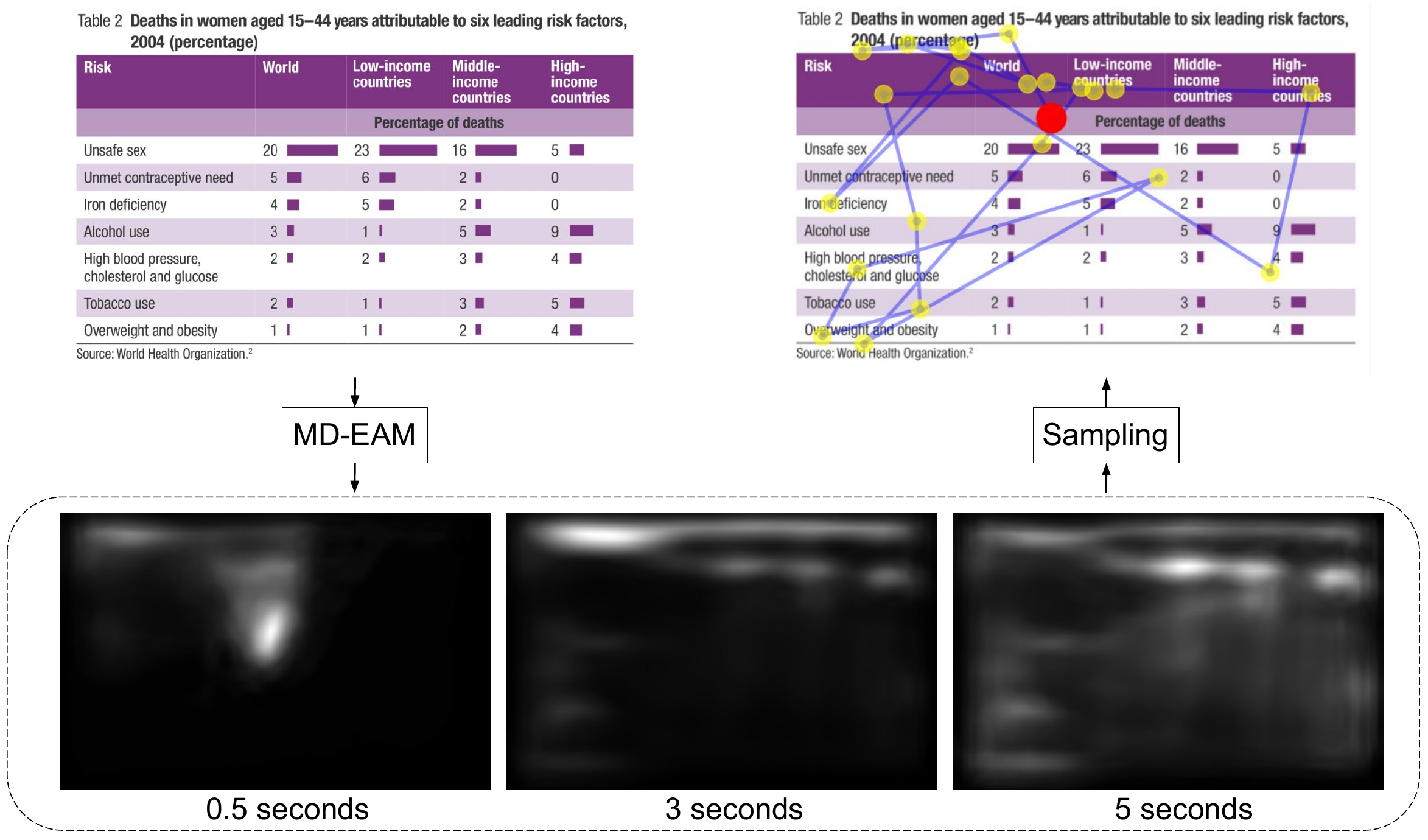}
    \caption{Our \methodNameShort method can predict human gaze scanpaths, that is, sequences of fixations, on information visualisations. It uses the Multi-Duration Element Attention Model~(MD-EAM), a model for predicting multi-duration element-level human attention maps, followed by a probabilistic approach for sampling gaze locations across visualisation elements. It outperforms existing scanpath models~\cite{assens2017saltinet,assens2018pathgan,bao2020human} in Sequence Score and ScanMatch metrics, and is the preferred scanpath prediction model by visualisation experts from our user study.
    }
    \label{fig:teaser}
\end{figure*}

Despite the importance of human gaze in information visualisation, for example to study media quality~\cite{bylinskii2017learning} or visual decision-making~\cite{feit2020detecting}, existing approaches to quantify users' visual attention require special-purpose eye tracking equipment \cite{kurzhals2014evaluating}.
However, eye trackers may not always be available.
They have to be calibrated to each user prior to first use~\cite{Harezlak2014TowardsAE}, and accurate gaze estimation is limited to confined areas in front of the display~\cite{Feit2017TowardEG}.
A popular approach to overcome these limitations is to instead use computational models of visual attention that can predict attention distributions over an image, such as saliency maps, without the need for any eye tracking equipment~\cite{boccignone2019look}.
Saliency modelling has been widely studied in computer vision~\cite{itti1998model, cornia2016multi, bao2020human, droste2020unified}, but has also found applications in human-computer interaction (HCI), such as for visual analytics~\cite{kurzhals2014evaluating}, optimising webpage designs~\cite{pang2016directing}, and re-targeting and thumbnailing on graphic designs~\cite{bylinskii2017learning}.

Information visualisations are fundamentally different from natural images:
they usually contain more text (e.g. titles, axis labels or legends) as well as larger areas with uniform colour and little to no texture (e.g. in bar plots or pie charts)~\cite{matzen2017data}.
These differences have triggered research into saliency models that are specifically geared to information visualisations, such as element-level saliency prediction~\cite{gupta2018saliency}.
However, saliency models are fundamentally limited in that they cannot predict the temporal dynamics of gaze behaviour.

Scanpath prediction is the task of predicting the sequence of fixations on an image~\cite{wang2017scanpath, matzen2017data}.
In contrast to saliency modelling, scanpath prediction inherently captures the stochastic and dynamic characteristics of visual attention over time.
Due to the large variability of human gaze, accurately predicting scanpaths is profoundly challenging~\cite{boccignone2019look}.
Prior methods for scanpath prediction have focused on natural scenes consisting of people and objects~\cite{pirinen2018deep, yang2020predicting}, on webpages~\cite{pang2016directing} or on graphical user interfaces \cite{xu16_chi}.
To the best of our knowledge, scanpath prediction on information visualisations has not yet been explored.
Modelling scanpaths on information visualisations can provide insights into both the rich spatial and temporal dynamics of human attention over time. 
As such, in contrast to saliency modelling, scanpath prediction can therefore help us better understand human visual behaviour while observing and visually processing information visualisations. 
Moreover, future work will be able to use our spatio-temporal models of attention for visualisation quality evaluation~\cite{behrisch2018quality} or visualisation optimisation~\cite{fosco2020predicting} without the need for any tedious and time-consuming eye tracking studies.
Scanpath prediction methods on information visualisations can be utilised as a tool to simulate human attention, which allow user models and simulations of visual attention on visualisations without the need for eye tracking equipment.

Since there is currently limited work understanding gaze behaviour on visualisations, we fill this gap and lay the foundations for a new line of research on scanpath prediction on information visualisations.
Inspired by similar investigations on natural images~\cite{fosco2020much}, we first conduct a systematic analysis of human gaze on visualisations from the widely used Massachusetts Massive Visualization Dataset (MASSVIS)~\cite{borkin2015beyond}.
Specifically, we analyse static and dynamic fixation density both across different visualisation elements -- such as title, data, axes, or labels -- as well as across viewers.
We find that title and graphical elements receive a significant amount of attention, particularly at the onset of the visual inspection process.
Afterwards, attention shifts to other textual elements, such as labels, followed by data-related components, such as annotations, legends or axes.
Moreover, attention towards objects and data elements is stable across time.

Informed by these findings, we propose \methodNameLong~-- a method to predict saliency and scanpaths on information visualisations.
The first stage of our method is the \methodSaliencyNameLong, which is a novel approach to predict multi-duration element-level human attention maps under multiple viewing durations.
The second stage of our method samples scanpaths from the multi-duration element-level human attention maps in a probabilistic way.
Through extensive evaluations on MASSVIS, we show that the novel element-wise attention maps and the data-driven sampling strategy allow our method to generate scanpaths of significantly better quality than previous methods. Moreover, they consistently outperform state-of-the-art methods with respect to several widely-used scanpath evaluation metrics. 
Our method achieves a relative improvement of 11.5\,\% in the Sequence Score~\cite{yang2020predicting}, and is best for the direction and position dimensions of MultiMatch~\cite{jarodzka2010vector}.
In addition, our method establishes a new state-of-the-art performance on the closely linked saliency prediction task on MASSVIS.
For example, it reaches a relative improvement of 23.6\,\% in the Pearson correlation coefficient under a 3-second viewing duration.

The contributions of our work are twofold.
First, we present a systematic analysis of gaze dynamics on visualisation elements and reveal both consistencies across visualisations and viewers as well as structural differences between different visualisation elements.
Second, we propose \methodNameLong, the first unified method for predicting multi-duration visual saliency and scanpaths on information visualisations.
Through extensive evaluations and a user study, we validate the effectiveness of our method, and report several fundamental findings of current scanpath metrics.

\section{Related Work}\label{sec:relatedwork}

Our work is related to previous works on (1) eye tracking for information visualisations, as well as to computational models for (2) visual saliency and (3) scanpath prediction. 
\subsection{Eye Tracking for Information Visualisations}

Eye tracking is widely used in information visualisations and visual analytics~\cite{kurzhals2014evaluating, burch2017eye}, given that eye gaze provides rich information about visual search and visual decision-making.
For instance, \citet{borkin2015beyond} assessed the key characteristics necessary to make visualisations recognisable.
Some other literature have proposed eye-tracking based visual analytics approaches, such as word-sized visualisations~\cite{beck2015word} and  under interactive visualisations~\cite{nguyen2015interactive}.
These works demonstrated the importance of eye tracking as a means to better understand gaze behaviour while viewing static as well as a component of visual analytics tools for dynamic information visualisations. 
However, while eye trackers have become cheaper and more readily available, they are still far from being pervasive, and have to be calibrated to each user before first use~\cite{Harezlak2014TowardsAE}, and often suffer from inaccuracies in everyday settings~\cite{li2019training}.

\subsection{Computational Modelling of Visual Attention}

Another line of work have addressed the limitation of eye-tracking equipment by proposing computational attention models.
Visual attention modelling, also known as saliency modelling, is a highly active research area in computer vision.
\citet{itti1998model} proposed one of the first bottom-up-models, that is, models that only consider visual features from a scene or image.
Since then, with large-scale annotated data from natural scenes becoming more easily available~\cite{xu2015turkergaze, jiang2015salicon}, several works have shown significant improvements in visual attention modelling~\cite{jiang2015salicon, kummerer2017deepgaze, cornia2018predicting}.
Multi-Duration Saliency Excited Model~(MD-SEM), a method to capture attention at multiple viewing durations~\cite{fosco2020much}, is the first method to provide insights into how human attention changes over time. 
It bridges statistical-level saliency and individual-level scanpath.
However, MD-SEM was proposed for natural images.
Therefore, we first have to test the performance of it on information visualisations.

Saliency models are not only useful to model human visual attention on natural scenes but also more broadly applicable, such as to information visualisations~\cite{matzen2017data}, web pages~\cite{shen2014webpage, feit2020detecting}, mobile user interfaces~\cite{gupta2018saliency, leiva2020understanding}, or graphical user interfaces \cite{xu16_chi}.
An increasing number of works have explored attention models in the context of information visualisations~\cite{7864468, matzen2017data}.
\citet{matzen2017data} proposed the data visualisation saliency (DVS) model that integrates bottom-up saliency maps of the Itti-Koch~\cite{itti1998model} model with text-region maps.
In follow-up work, the same authors showed that attention towards outliers in data visualisations is heavily influenced by the task~\cite{matzen2020task}.
Complementing the notion of saliency, others have proposed visual importance as a concept to model the level of importance of different visualisation elements~\cite{o2014learning, bylinskii2017learning}.
\citet{fosco2020predicting} proposed the Unified Model of Saliency and Importance~(UMSI) -- a method to predict importance maps across five types of graphic designs, including infographics, movie posters, mobile user interfaces, advertisements and webpages.

\subsection{Scanpath Prediction}
Models of visual attention only provide aggregated statistics, which has triggered research into the complementary task of scanpath prediction, that is, the task of predicting a sequence of fixations over a visual stimulus~\cite{itti1998model}.
Scanpath prediction has been studied on different types of visual stimuli such as natural scenes~\cite{coutrot2018scanpath, verma2019hmm, xia2019predicting, yang2020predicting}, virtual reality environments~\cite{hu2020dgaze, hu21_tvcg}, and graphical layouts~\cite{jokinen2020adaptive}.
Scanpath prediction is even more challenging given that fixation locations vary a lot across viewers~\cite{assens2017saltinet}.
Early work on scanpath prediction has typically used bottom-up saliency maps to predict gaze shifts~\cite{ brockmann2000ecology, boccignone2010gaze}. 
Other models have incorporated cognitively plausible mechanisms, such as inhibition of return~\cite{itti1998model, zanca2019gravitational, sun2019visual} or foveal-peripheral saliency~\cite{wang2017scanpath, wloka2018active, bao2020human}.
\citet{boccignone2019look} have created a three-stage processing model with a centre-bias, a context/layout and an object-based model to predict scanpaths on natural scenes.
Scanpath prediction under object detection~\cite{mathe2016reinforcement, pirinen2018deep}, visual search~\cite{yang2020predicting}, or visual question-answering~\cite{chen2021predicting} is solved by reinforcement learning.
\citet{islam2020st} have proposed a multitask-learning framework for segmentation and scanpath prediction and showed that this approach can take advantage of a segmentation task.
HMM-based scanpath prediction methods either split an image into several grids and regard each grid as a single state of observation~\cite{verma2019hmm}, or classify the fixations into several states~\cite{coutrot2018scanpath}.

Large-scale datasets~\cite{xu2015turkergaze, jiang2015salicon} have paved the way for the use of deep learning methods for scanpath prediction on natural images.
Saltinet~\cite{assens2017saltinet} has extended saliency maps to saliency volumes, from which sample scanpaths were created. 
\citet{kummerer2022deepgaze} proposed the DeepGaze III model that allowed them to predict next fixations from saliency maps and previous scanpaths.
PathGAN~\cite{assens2018pathgan} was the first end-to-end model that relied on a generative adversarial network (GAN) for scanpath prediction. 
It combined a VGG network~\cite{Simonyan15} to encode the image with an LSTM-based generator to predict scanpaths as well as a discriminator to distinguish the generated scanpaths from the real ones.
Since insufficient gaze data are collected on visualisations, not surprisingly, no deep learning-based scanpath prediction model is designed for information visualisations. Therefore, it is essential to understand gaze behaviour on information visualisations and apply key findings to our model to alleviate the data scarcity problem.

\section{Analysing Gaze Behaviour on Information Visualisations}\label{sec:analysis}

Although eye tracking has been widely used in information visualisation research, the ways in which viewers look at visualisations remain under-explored.
While several works have investigated eye movements on visualisations~\cite{borkin2015beyond,matzen2017patterns}, they have been limited to statistical results, rather far from revealing gaze dynamics.
To shed more light on gaze dynamics while viewing information visualisations and to inform the design of our method for scanpath prediction, we conducted fundamental analyses on the Massachusetts Massive Visualization Dataset (MASSVIS).

\subsection{The MASSVIS Dataset}
\label{ss:Dataset}

MASSVIS~\cite{borkin2013makes, borkin2015beyond} consists of more than 5,000 static information visualisations and, as such, is one of the largest and most widely used datasets.
It covers various types of visualisations, such as government reports, infographic blogs, news media websites, and scientific journals, and provides detailed annotations of visual elements, such as titles, data, axes and legends.
The dataset also provides gaze data recorded from human viewers for a subset of 393 visualisations.
Gaze data was collected during a memorability task that involved two phases:
in the encoding phase, viewers were given 10 seconds to memorise each visualisation.
In the following recognition phase, viewers were asked to recognise the visualisation within two seconds.
Given that the visualisations were shown only 2 seconds for the recognition stage, we only analysed visualisations and gaze data in the encoding stage.
The gaze data from the encoding stage were collected from 33 viewers and 16.7 viewers per visualisation. 
The mean scanpath length on this data was 37.4 fixations~($\sigma$\,=\,6.64) with a maximum of 55 fixations and the mean duration of 219.17\,ms~(see Figure~1 in supplementary material for fixation duration distribution).
The element taxonomy and annotations were derived from MASSVIS~\cite{borkin2015beyond}.
The first fixations were discarded in every scanpath due to the experimental setting, where a fixation cross showed up right before the image appeared on the screen~\cite{borkin2015beyond}~(see Figure~2 in supplementary material).
Regarding fixations that landed on white spaces, we followed a similar procedure to prior work~\cite{wang2022impact} and removed 12.2\% of total fixations that did not land on any visual elements.

\begin{figure}[!t]
  \centering
    \includegraphics[width=\linewidth]{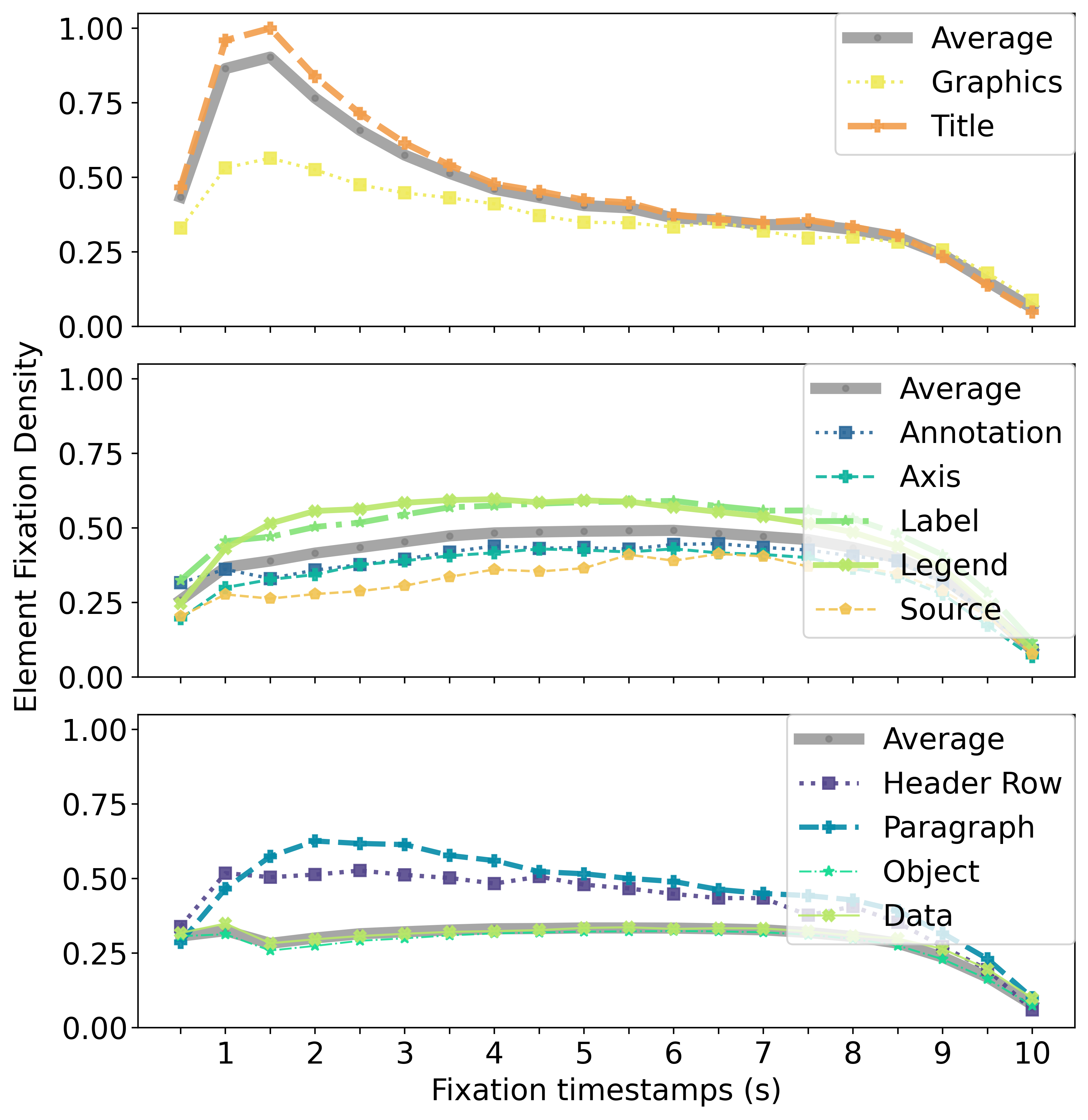}
  \caption{
    Element fixation density \yao{of visual elements over time} on the MASSVIS dataset \yao{in a 0.5-second bin}. Title and Graphics draw a substantial amount of attention in the beginning~(top), then attention shifts to other textual elements (Label and Source etc.), and data-related elements, (Annotation, Legend and Axis)~(middle). Meanwhile, attention towards Object and Data is consistent across time~(bottom).} %
\label{fig:analysis_generaltrend}
\end{figure}

\begin{figure*}[!t]
\centering
     \includegraphics[width=0.9\linewidth]{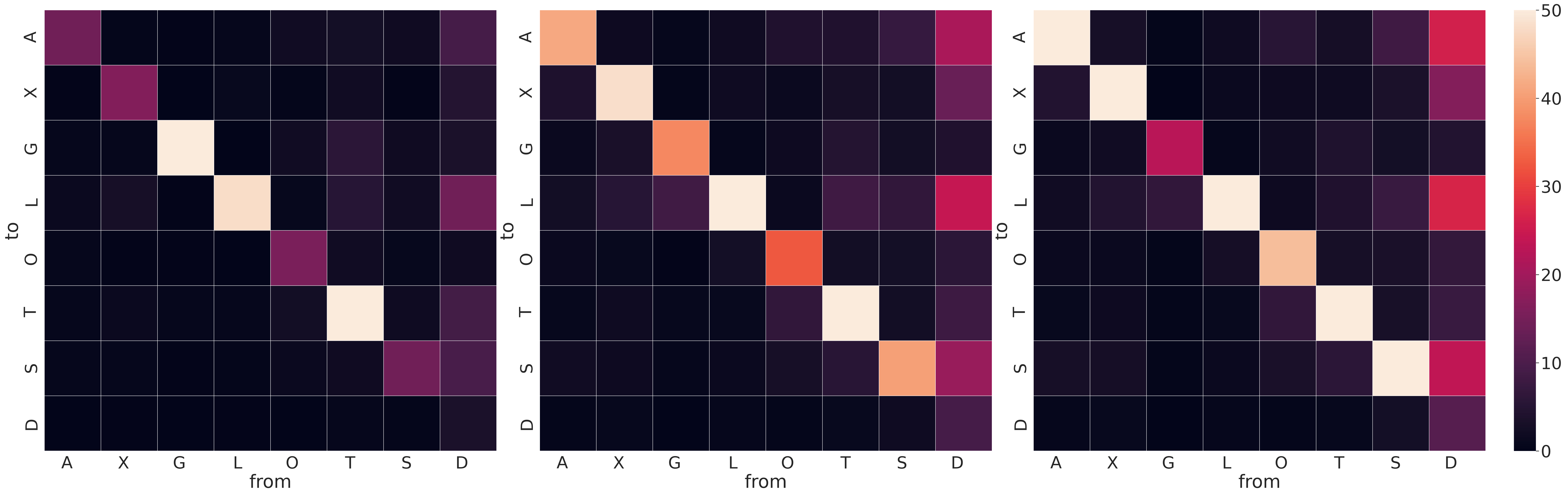}
     \caption{Human gaze transition matrices under three different viewing durations. Left: First 3 seconds. Middle: From 3 to 5 seconds. Right: From 5 to 10 seconds. Viewers tend to look at Title and Legend continuously before jumping to other regions, while they tend to read Data in cooperation with Annotation, Axis, Legend and Source etc.. A: Annotation, X: Axis, G: Graphics, L: Legend, O: Object, T: Title, S: Source etc., D: Data.}
  \label{fig:analysis_transitionmatrics}
\end{figure*}

\subsection{Fixation Density on Visual Elements}
\label{ss:EFD}

Compared to natural images, information visualisations often contain larger areas with uniform colours as well as small, yet important, areas such as text~\cite{matzen2017data}.
It is therefore conceivable that, in addition to their information content, the relative saliency of individual visual elements influences if and when they are being looked at during the execution of a scanpath.
It currently remains unclear, however, how salient different elements are overall as well as relative to each other.
Furthermore, it is also unknown whether human attention is evenly distributed over a particular element over time or whether it changes as a function of when the visualisation element is more attractive.
In the theory of visual world paradigm~\cite{huettig2005word, salverda2017visual}, the proportion of fixations on each target is plotted over time to show how visual attention shifts to different items in a scene during the comprehension of spoken language. However, it requires each target to have a similar size, making it not applicable in information visualisations.

Inspired by this, we propose the Element Fixation Density~(EFD) measure to quantify how visual attention evolves in arbitrary size of visual elements. EFD is defined as the accumulated number of gaze fixations divided by the covering area of fixation targets~\cite{shojaeizadeh2016density}. Derived from the term ``Fixation Density'' introduced by~\cite{shojaeizadeh2016density}, the fixation target in EFD is set to the sum area of one kind of visualisation element, such as title, data, and legend.
Inspired by previous attention dynamics analysis \cite{fosco2020much}, we used k-means to cluster the elements that have similar attention dynamics.
\yao{
The distance of two elements $x$ and $y$ is calculated as
$d_{xy} = \sum\limits_{t} \lvert EFD_x(t) - EFD_y(t)\rvert$, $t$ for every 0.5 seconds.
\autoref{fig:analysis_generaltrend} shows the EFD of visual elements over time in a 0.5-second bin ($x$-axis, 0\,--\,10\,s), clustered into three groups by the k-means algorithm.
}
As can be seen from the figure, \textit{Title} and \textit{Graphics} draw a substantial amount of attention in the beginning, then attention shifts to other textual elements (\textit{Label} and \textit{Source etc.}), and Data-related elements (\textit{Annotation}, \textit{Legend} and \textit{Axis}). Meanwhile, attention towards \textit{Object} and \textit{Data} is consistent across time.
See Figure~4 in supplementary material for example visualisations with annotated semantic regions.
In the following paragraphs, we discuss when an element attracts attention in visualisations in detail.

\textit{Text~(\textit{Title}, \textit{Source etc.}, \textit{Paragraph}, and \textit{Label}).}
Previous work reported the bias of human attention towards text regions~\cite{matzen2017data} but did not reveal \yao{the temporal preference of text elements}.
\autoref{fig:analysis_generaltrend} shows that most text elements~(\textit{Title}, \textit{Paragraph}, \textit{Label}) receive a large EFD. 
For text categories that are not directly related to data, such as \textit{Title} and \textit{Paragraph}, the attention first increases but then reaches a peak at 0.5\,--\,2.5\,s. 
This suggests that viewers tend to examine these regions at the very beginning of observation, which is in line with previous analyses on the time to first fixation of different elements~\cite{bylinskii2015eye}. 
Then, the interest in these elements decreases afterwards, especially for \textit{Title}. Data-related text elements such as \textit{Label} and \textit{Source etc.} reach the peak around 5.5\,--\,7s. The highest EFD across all elements appears in \textit{Label}.

\textit{\textit{Data} and Data-related Elements.}
\autoref{fig:analysis_generaltrend} shows that data-related elements~(\textit{Legend}, \textit{Annotation} and \textit{Axis}) have lower EFD than \textit{Legend}, while the interest towards \textit{Legend} is as great as for \textit{Title} after 4\,s.
\textit{Data} areas cover more than half of all pixels in visualisations~\cite{borkin2015beyond} but their EFDs are the lowest among all elements. 
The attention towards \textit{Data} decreases over 1\,--\,2\,s, then gradually increases.
This pattern also appears in data-related elements, and we notice the interest stays undiminished for an extended period.
\textit{Legend} reaches its peak around 2.5\,s, and it stays at a high level of EFD until 6\,s.
Attention towards \textit{Annotation} and \textit{Axis} starts to grow at 4\,s, and remains at a high level until 7\,s. 
We find the peak of \textit{Data} occurs around 6\,s, which agrees with the trend of data-related elements.
These findings suggest that viewers usually examine the \textit{Title} in first glances, then pay attention to data-related elements.
Around 5\,--\,7\,s, viewers tend to observe visualisations by alternating between \textit{Data} and descriptive elements.

\textit{Object}.
Objects are ``either realistic photographs or abstract drawings or pictograms that can be recognised by human''~\cite{borkin2015beyond}.
Object persistence is a well-known recognition process~\cite{scholl2007object}.
We find that attention density within \textit{Object} is comparatively low in memorability tasks. Even though \textit{Object} takes 7.67\,\% of image space pixel-wise, the fixations make up only 2.16\,\%. The attention pattern towards \textit{Object} is very similar to \textit{Data}, which reaches the lowest EFD at 1\,--\,1.5\,s and then peaks at 5\,--7\,s.
We suggest this pattern may be caused by the well-known Inhibition of Return~(IOR)~\cite{itti1998model}. Since \textit{Object} contains relatively limited information compared to textual elements, viewers tend to postpone their attention towards the entire \textit{Object} regions for a later time. After the effective period of IOR, the interest towards \textit{Object} increases again.

\subsection{Attention Dynamics for Individual Viewers}
\label{ss:individuallevelattentionshift}

Our analyses so far focused on the temporal dynamics of gaze on visualisations across all viewers.
However, it is well-known that, in general, gaze behaviour contains not only person-independent but also person-specific information~\cite{zhang2018training}.
We therefore analysed the individual scanpath trends of 10 viewers in MASSVIS, where all viewers observed at least 75\% of all 393 visualisations.
In \autoref{ss:EFD}, we reported attention dynamics patterns for every kind of element.
For a better understanding of attention dynamics, we merged the four text elements that have the same dynamic patterns into one. The final eight types of visual elements are A: \textit{Annotation}, X: \textit{Axis}, G: \textit{Graphics}, L: \textit{Legend}, O: \textit{Object}, T: \textit{Title}, S: \textit{Source, paragraph, label, and header row text, denoted as Source etc.}, D: \textit{Data}.

\begin{figure}[!t]
  \centering
    \includegraphics[width=\linewidth]{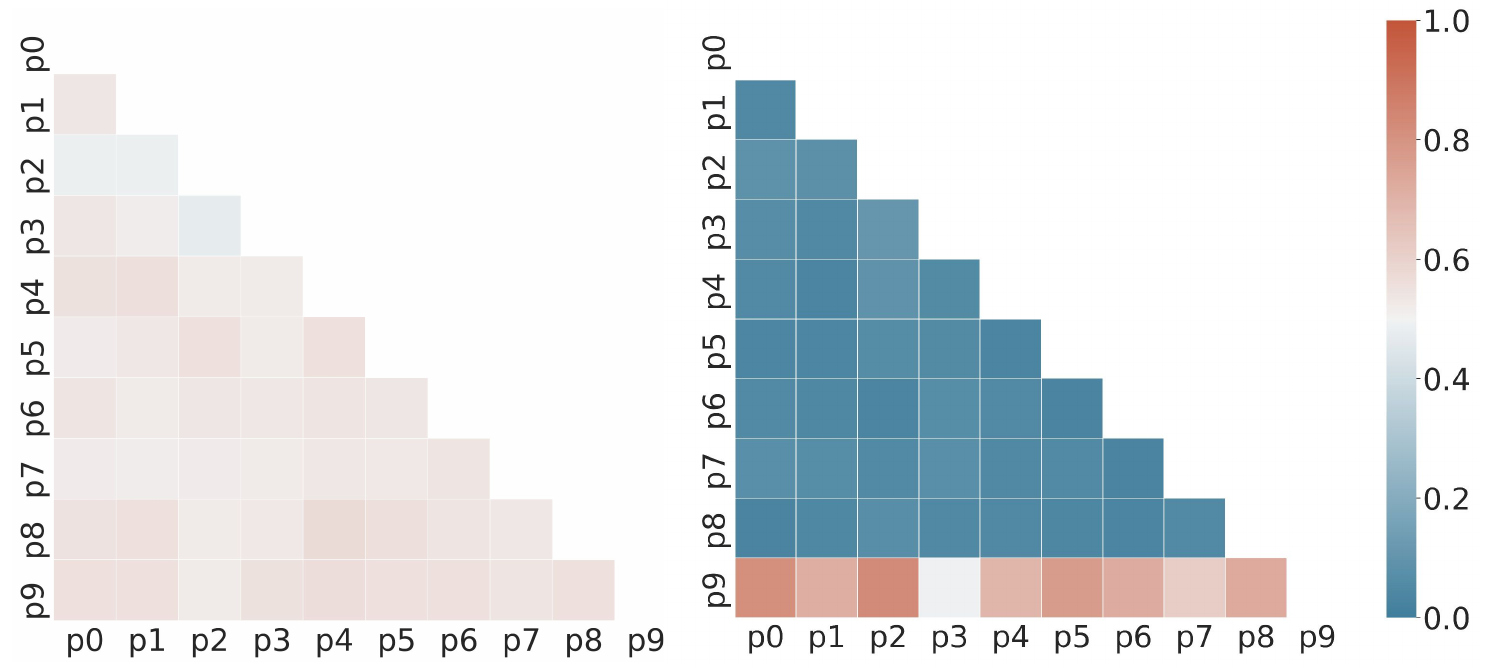}
  \caption{Two statistical results of attention dynamics across viewers in MASSVIS~\cite{borkin2015beyond}. Left: Sequence Score~\cite{yang2020predicting} of scanpaths. Right: Kullback-Leibler divergence of human gaze transition matrices. Individual participants are denoted as p0\,--\,p9. It shows substantial similarities in fixation distributions between most participants (right), but the scanpaths vary a lot from each other (left).}
  \label{fig:analysis_fixationdensity}
\end{figure}

\textit{Sequence Score.}
We reported that attention dynamics towards elements are consistent across visualisations and viewers, but the individual-level analysis is the key to understanding scanpaths. Therefore, we converted a scanpath to a sequence of letters by assigning each fixation to a unique letter based on the element at which it was drawn.
We used the Sequence Score~\cite{yang2020predicting} to quantitatively examine how similar scanpaths are within viewers. To compute the Sequence Score, the Needleman-Wunsch algorithm~\cite{needleman1970general} was used to calculate the minimum number of operations needed to change one string into another. Each mismatch or gap between two strings penalises the final score. We observed a low similarity of Sequence Score within viewers in Figure~\ref{fig:analysis_fixationdensity}, left, which means \textit{different viewers observe the same visualisation in quite different ways}. Moreover, the Sequence Score within the first 5 seconds was also calculated. However, to our surprise, the Sequence Score within the first 5 seconds was even slightly lower than for the entire 10 seconds. This may suggest that the attention dynamics in the early observation period are more unstable than in the late observation period.

\textit{Transition Matrix.}
To give a panoptic view of individual attention dynamics on visualisations across images, we adopted the concept of transition matrix from Hidden Markov Models to describe gaze shifts.
We computed the transition matrix of fixations in scanpath strings. Each letter in scanpaths was considered a hidden state, and changes between neighbour letters were state transitions. The average transition matrices across all viewers within three different durations are demonstrated in Figure~\ref{fig:analysis_transitionmatrics}, that is, before 3\,s, 3\,s to 5\,s and from 5\,s to 10\,s. The diagonal values of the transition matrices stand for self-transition, which means the next fixation stays in the same kind of element as the previous fixation. The highest self-transition appeared in \textit{Legend}~(L), while \textit{Title}~(T) comes second. It indicates that people tend to keep reading legends and titles before jumping to other regions. The lowest self-transition appears in \textit{Objects}~(O) and \textit{Data}~(D). It indicates that people alternately read these regions or only glance at these regions rather than focus on them. We also found some consistent attention dynamics for elements. The gaze shifts from the \textit{Data} are more likely shifting to \textit{Annotation}~(A), \textit{Axis}~(X), \textit{Legend}~(L) and \textit{Source etc.}~(S). The transitions from X to L, from T to G, and T to L are also relatively high. We also observe consistent attention dynamics across viewers under transition matrices.
To quantify the similarity of individual dynamics, we computed the Kullback-Leibler divergence~(KL) within ten viewers (see Figure~\ref{fig:analysis_fixationdensity}, right). The lowest KL of 0.023 and the highest KL of 0.827 demonstrated substantial similarities of attention dynamics across individuals. It suggests that \textit{the individual-level attention dynamics of viewing information visualisations are consistent with those on the element level}.

\section{\methodNameLong}\label{sec:method}
\begin{figure*}[ht]
    \centering
    \includegraphics[width=\linewidth]{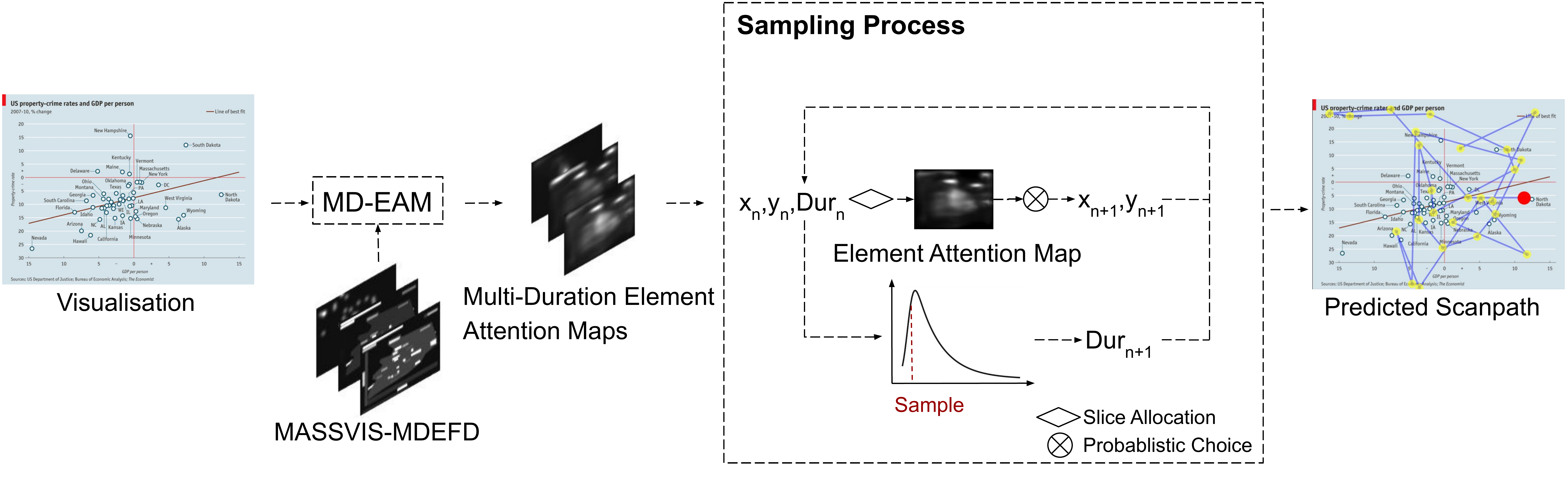}
    \caption{Overview of our method for probabilistic scanpath prediction on information visualisations. Multi-Duration Element Attention Model (MD-EAM) is fine-tuned by the MASSVIS Multi-Duration Element Fixation Density~(MASSVIS-MDEFD) dataset, and generates multi-duration element attention maps for saliency prediction. The duration is sampled from the ex-Gaussian distribution estimated from the MASSVIS training set. The Element Attention Map is selected by timestamp. Then, fixations are sequentially sampled from the selected map by probabilistic choice.}
    \label{fig:method_overview}
\end{figure*}

Our analyses yielded several insights that are important when designing a method for predicting scanpaths on information visualisations. 
We found that \textit{Title} and \textit{Graphics} receive a significant amount of attention, particularly at the onset of the visual inspection process.
Afterwards, attention shifts to other textual elements~(\textit{Label} and \textit{Source}), followed by data-related components~(\textit{Annotation}, \textit{Legend} and \textit{Axis}).
Moreover, attention towards \textit{Object} and \textit{Data} is consistent at a stable level across time.
Specifically, we found that though gaze patterns across viewers are highly consistent, individual scanpaths show significant variability. 
Taken together, these characteristics render the task of scanpath prediction particularly challenging.
We therefore designed our \methodNameLong~with the specific goal of preserving this stochastic nature of fixations within a scanpath.
Our method combines two original contributions towards this goal:
a \methodSaliencyNameLong that builds on the architecture of MD-SEM~\cite{fosco2020much} but better preserves element-level spatial information, as well as a probabilistic approach to sample scanpaths from these attention maps.
\autoref{fig:method_overview} gives an overview of our method.

\subsection{\methodSaliencyNameLong}
\label{ss:mdeam}

Our analyses showed that attention dynamics on visualisation elements are large, which indicated that different elements are salient under different durations.
From our analysis, we found out that where viewers tend to focus on a visualisation depends on how long they have been observing it (see~\autoref{fig:analysis_generaltrend}). 
Thus, a single saliency map is not representative enough to describe the gaze dynamic over time. 
MD-SEM~\cite{fosco2020much} is the first and currently state-of-the-art method to model multi-duration saliency, that is, a model that can predict saliency maps for different viewing durations.
The model learns temporal attention dynamics using a three-branch weight-sharing network, and predicts the attention distribution for a certain duration in each branch. 
From our perspective, there are two main drawbacks of MD-SEM: 1) Saliency dispersion to white spaces; and 2) lack of spatial information, such as element bounding boxes.

Thus, we leverage the above drawbacks by fine-tuning MD-SEM on element fixation density maps.
As \autoref{ss:EFD} defines, the EFD of an element is calculated by the accumulated fixations divided by the element area. We assign the element EFD as the uniform value to all pixels in that element, and truncate fixations to three continuous observation periods~(e.g. 0\,--\,0.5\,s, 0.5\,--\,3\,s, and 3\,--\,5\,s).
We denote these EFD maps as MASSVIS Multi-Duration Element Fixation Density~(MASSVIS-MDEFD), and the fine-tuned MD-SEM model as \methodSaliencyNameShort.
Thus, we leverage the above drawbacks by fine-tuning MD-SEM on the MASSVIS-MDEFD dataset.
\methodSaliencyNameShort shows better capacity in preserving element-level attention distribution~(see Figure~5 in supplementary material).

\subsection{Probabilistic Scanpath Sampling}

Previous work has reported that recurrent layer-based networks regressed to the image centre in scanpath prediction~\cite{assens2017saltinet}, which also occurred on information visualisations.
To tackle the centre-regress problem, we propose a probabilistic sampling method to generate realistic scanpaths.

\textit{Duration Prediction.}
Previous literature~\cite{staub2013individual} reported that the fixation duration is stimuli-dependent, and is close to the exponentially modified normal distribution~(ex-Gaussian). In our method, we first estimate the ex-Gaussian parameters from training data, and sample durations from the distribution. We follow this strategy to estimate the three parameters, $\mu$, $\delta$, and $\tau$ for the ex-Gaussian distribution.

\textit{Slice Allocation.}
The scanpath length and durations were sampled from the distribution of the training data~\cite{assens2017saltinet}, while the number of fixations in each slice of the attention map of \methodSaliencyNameShort is based on fixation timestamps. 
As shown in~\autoref{fig:method_overview}, the probability-based algorithm randomly samples fixations from multi-duration element attention maps.
With prior knowledge of the length and duration of the scanpath, we can easily decide how many fixations are in each slice of the attention map.
Inspired by Saltinet~\cite{assens2017saltinet}, each slice of attention maps is regarded as a probability distribution, and the first position $X_0$ in each slice is randomly sampled from the attention map.

To mimic gaze shift, we create a foveal mask $M_n$ by multiplying the allocated slice of the attention map with a Gaussian kernel centred at the fixation position $X_{n}$.
Then, the next fixation position $X_{n+1}$ stays in the foveal region of $M_n$~(see Algorithm~\ref{alg:FPASA}). %
This process will continue multiple times in each slice of the attention map. The final scanpath is generated by concatenating fixations from all slices of attention maps.

\begin{algorithm}
\begin{algorithmic}[1]
\Procedure{FixationSampling}{$X_n, M_n$}
\State $dur=SampleFixationDuration(\mu, \delta, \tau)$
\Comment{Randomly sample a duration from the ex-Gaussian distribution}
\While{current timestamp in range}
\State Find the current slice $AttMap$
\State $X_{n+1}=ProbablisticChoice(AttMap \cdot M_n)$
\State $M_{n+1}=Gaussian(X_{n+1})$
\State \textbf{return} $X_{n+1}, M_{n+1}$
\EndWhile
\EndProcedure
\end{algorithmic}
\caption{Foveal Attention Shift Algorithm}\label{alg:FPASA}
\end{algorithm}

\section{Experiments}\label{sec:experiments}

We carried out a series of experiments to compare the performance of \methodNameShort with state-of-the-art saliency and scanpath prediction methods. Different ablated versions of the method itself were also evaluated.

\begin{table*}[ht]
    \caption{Quantitative evaluation on MASSVIS for a 5-second ground truth in terms of Sequence Score~(SS), Scanmatch, scaled Time Dimension Embedding~(sTDE), Dynamic Time Warping~(DTW) and MultiMatch metrics~(shp: shape, dir: direction, len: length, pos: position, dur: duration). Best results are shown in \textbf{bold}, second best are \underline{underlined}. Stars indicate statistical significance of the difference between Ours and MD-SEM (**: p\,<\,.01; ***: p\,<\,.001).}
    \label{table:scanpath_5s}
    \centering
    \begin{tabular}{lllllllllllllll}
    \toprule
    \multirow{2}{*}{\textbf{Methods}} & \multicolumn{2}{c}{\textbf{SS~$\uparrow$}} & \multicolumn{2}{c}{\textbf{ScanMatch~$\uparrow$}} & \multicolumn{2}{c}{\textbf{sTDE~$\uparrow$}} & \multicolumn{2}{c}{\textbf{DTW~(2D)~$\downarrow$}} & \multicolumn{5}{c}{\textbf{MultiMatch}~$\uparrow$}\\
    & \textit{mean} & \textit{best} & \textit{mean} & \textit{best} & \textit{mean} & \textit{best} & \textit{mean} & \textit{best} & \textbf{shp} & \textbf{dir} & \textbf{len} & \textbf{pos} & \textbf{dur} \\
    \midrule
    Human & 0.584 & 0.651$^\dag$ 
    & 0.532 & 0.645$^\dag$
    & 0.924 & 0.943$^\dag$ 
    & 5311.23 & 3433.68$^\dag$ 
    & 0.958 & 0.800 & 0.952 & 0.818 & 0.730\\
    \midrule
    PathGAN~\cite{assens2018pathgan} & 0.390 & 0.503 
    & 0.232 & 0.255
    & \textbf{0.910} & \textbf{0.937} 
    & 6840.86 & \underline{4495.89} 
    & \textbf{0.974} & 0.671 & \textbf{0.964} & \underline{0.767} & 0.691\\
    DCSM~\cite{bao2020human} & \underline{0.400} & 0.580 
    & 0.328 & \underline{0.458}
    & 0.879 & 0.908 
    & \textbf{6395.57} & \textbf{4292.44} 
    & 0.924 & \underline{0.724} & 0.902 & 0.756 & \textbf{0.755}\\
    Saltinet~\cite{assens2017saltinet} & 0.388 & \underline{0.676} 
    & \underline{0.331} & 0.451
    & 0.875 & 0.876 
    & 12758.51 & 10546.33 
    & 0.887 & 0.689 & 0.842 & 0.684 & 0.708\\
    \methodNameShort~(ours) & \textbf{0.446$^{***}$} & \textbf{0.724$^{**}$} 
    & \textbf{0.387$^{***}$} & \textbf{0.503$^{***}$}
    & \underline{0.906} & \underline{0.925} 
    & \underline{6529.11} & 4683.44 
    & \underline{0.943} & \textbf{0.728} & \underline{0.935} & \textbf{0.771} & \underline{0.712}\\
    \bottomrule
    \end{tabular}\\
    \vspace{2pt}
    \footnotesize{$^\dag$ Scanpaths are not compared with themselves}
    \label{table:quantitative}
\end{table*}

\subsection{Dataset}
\label{E:Dataset}
Since the provided fixations in the SALICON dataset~\cite{jiang2015salicon} lacked timestamps, we retrieved fixation duration by applying the IDT~(Identification by Dispersion Threshold) algorithm~\cite{komogortsev2010standardization} on raw gaze data to prepare the SALICON-MD~(Multi-Duration) dataset.
We truncated fixations in MASSVIS~\cite{borkin2013makes, borkin2015beyond} to the first 5 seconds to make fair comparisons with baseline methods.
MASSVIS-MD~(Multi-Duration) is a dataset created according to the following gaze timestamps: 0\,--\,0.5\,s, 0.5\,--\,3\,s, and 3\,--\,5\,s.
We used this dataset to fine-tune MD-SEM~\cite{fosco2020much} on information visualisations as a baseline. 
Then, we prepared MASSVIS-MDEFD from the element annotations of MASSVIS~\cite{borkin2015beyond} with the same durations as MASSVIS-MD for training \methodSaliencyNameShort.
To validate the generalisability of our model, we wanted to created train and evaluation sets that are balanced w.r.t. the source and visualisation type. Based on the naming convention of each sample from the MASSVIS dataset, we sorted all the files by name and selected every sixth file for the evaluation set.
This assures the evaluation set is balanced by both source and visualisation type.
More details are available in Table 2 in supplementary material.
All evaluations on MASSVIS followed the same split policy.

\subsection{Implementation Details \& Model Training}
The MD-EAM model was fine-tuned on MASSVIS-MDEFD for 6 epochs starting from the official CodeCharts1K weights of MD-SEM~\cite{fosco2020much}.
We preserved the original saliency maps at 0.5\,s duration to supervise the \methodSaliencyNameShort branch to align to the centre bias phenomenon that appeared in the first fixations of human gaze data~(see Figure~2 in supplementary material). For the other two branches of MD-EAM, we employed the MASSVIS Multi-Duration Element dataset at 3\,s and 5\,s.
For duration estimation, the parameters of the ex-Gaussian distribution were computed as $\mu$\,=\,124.06, $\delta$\,=\,17.49, and $\tau$\,=\,89.37. All experiments were conducted on a single NVIDIA Tesla V100 GPU with 32\,GB VRAM. See supplementary material for the training details.

For saliency prediction, we fine-tuned the MD-SEM model on MASSVIS-MD for six epochs starting from the official CodeCharts1K~\cite{fosco2020much} weights. DVS~\cite{matzen2017data} was used as-is to predict saliency maps on MASSVIS.
For scanpath prediction, we trained PathGAN~\cite{assens2018pathgan} on SALICON~\cite{jiang2015salicon} and fine-tuned it on the MASSVIS dataset~\cite{borkin2015beyond}.
We used the official implementation of Saltinet~\cite{assens2017saltinet} as-is to predict scanpaths on MASSVIS. For DCSM~\cite{bao2020human}, we were in contact with the corresponding author who sent us all the predicted saliency maps and scanpaths -- their codebase is not available publicly.

\subsection{Scanpath Prediction}
\label{E:scanpathprediction}

Since there is currently no scanpath prediction method for information visualisations, we compare our method to three state-of-the-art methods for natural scenes: DCSM~\cite{bao2020human}, PathGAN~\cite{assens2018pathgan} and Saltinet~\cite{assens2017saltinet}.

\textit{Metrics.}
Generated scanpaths were compared to human scanpaths using several evaluation metrics. 
We chose the five most currently used metrics to quantify the scanpath performance: Sequence Score~\cite{yang2020predicting}, Dynamic Time Warping~(DTW)~\cite{muller2007dynamic}, scaled Time-Delayed Embedding~(sTDE)~\cite{zanca2018fixatons, wang2011simulating}, ScanMatch~\cite{cristino2010scanmatch} and MultiMatch~\cite{jarodzka2010vector}. For Sequence Score, ScanMatch, DTW and sTDE, the \textit{mean} and \textit{best} evaluation scores were reported. While the \textit{mean} evaluation scores are the averages of all human and predicted scanpath pairs, the \textit{best} evaluation scores are the maximum of all pairs for each prediction~\cite{faggi2020wave, chen2021predicting}.

\begin{itemize}
    \item \textit{Dynamic Time Warping~(DTW).} DTW calculates an optimal match between two given sequences with specific rules, with smaller values indicating better performance~\cite{berndt1994using}. In this paper, we computed DTW in two-dimensional position coordinates.
    \item \textit{Scaled Time-Delay Embedding~(sTDE).} Time-delay embedding similarity refers to the inclusion of historical information in dynamic system models~\cite{pan2020structure}. It is a value between 0~(worse) and 1~(better).
    \item \textit{ScanMatch.} ScanMatch~\cite{cristino2010scanmatch} is a patch-based similarity approach inspired by the Needleman–Wunsch algorithm~\cite{needleman1970general}.
    It is a value between 0~(worse) and 1~(better). In this paper, we set no time bin for ScanMatch to ignore duration.
    \item \textit{Sequence Score~(SS).} The Sequence Score is normalized between 0 and 1. A detailed definition of Sequence Score can be found in \autoref{ss:individuallevelattentionshift}.
    \item \textit{MultiMatch.} MultiMatch~\cite{jarodzka2010vector} is a multidimensional vector-based approach. After the alignment of vector shapes, the length, position, direction, and duration of fixations are computed. All the obtained values are normalised between 0~(worse) and 1~(better).
\end{itemize}

\begin{figure*}[!t]
    \centering
     \includegraphics[width=\linewidth]{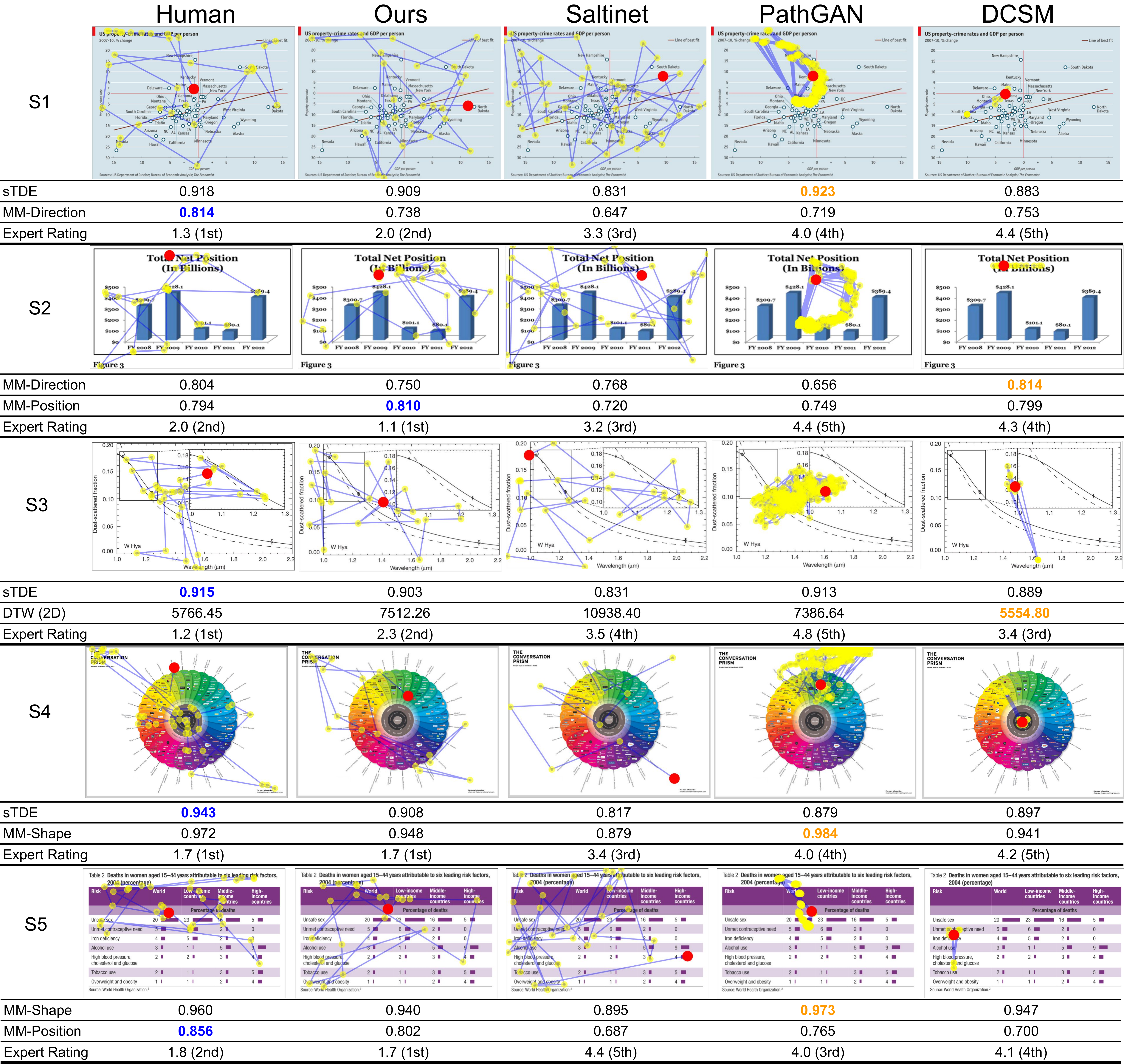}
    \caption{
    Examples of mismatches between scanpath prediction performance as seen through the evaluation metrics and visualisation expert ratings.
    Each row (one visualisation from MASSVIS) shows one metric that is contradictory to expert rating~(orange), and one metric that is consistent with expert rating~(blue). Our method and the human baseline have consistent metrics with expert rating. PathGAN and DCSM rank the highest in some metrics, even though the produced scanpaths were ranked much lower in our expert user evaluation. See Figure~6 in supplementary material for full table.}
    \label{fig:qualitative}
\end{figure*}

\textit{Results.}
\autoref{table:scanpath_5s} summarizes quantitative results on scanpath prediction for a 5-second ground truth. Metrics between real viewers on the same images are used as a golden standard of scanpath quantification, which is denoted as Human in \autoref{table:scanpath_5s}. 
Our method ranks first in Sequence Score, ScanMatch, MultiMatch-direction, and MultiMatch-position, and second in MultiMatch-shape, MultiMatch-length, and sTDE.
For DCSM, only one prediction for each visualisation is generated. For PathGAN, Saltinet and \methodNameShort, we generate the same number of predictions as human scanpaths for each visualisation~(16.7 per visualisation).
PathGAN and Saltinet are evaluated by conducting the Hungarian Algorithm~\cite{kuhn1955hungarian} with original setting, while our \methodNameShort is evaluated by averaging exhaustive matches between the generated scanpaths with human scanpaths. Quantitative results on scanpath prediction for the full 10-second ground truth can be found in supplementary material. Qualitative results are illustrated in \autoref{fig:qualitative}.

\subsection{Saliency Prediction}
\label{E:saliencyprediction}
We compare our saliency prediction results against the state-of-the-art DVS~\cite{matzen2017data} model on visualisations, and two on natural scenes~(MD-SEM~\cite{fosco2020much} and DCSM~\cite{bao2020human}). The MASSVIS-MDEFD that we created for training \methodSaliencyNameShort is also evaluated as a baseline.

\begin{table*}[t]
    \caption{Evaluation of saliency methods under 3-second and 5-second durations. Best results are shown in \textbf{bold}, second best are \underline{underlined}. The MASSVIS-MDEFD that we created for training \methodSaliencyNameShort is also evaluated as a baseline~(see Figure~5 in supplementary material). Stars indicate statistical significance of the difference between our Full Model and the best baseline model (**: p\,<\,.01; ***: p\,<\,.001).}
    \centering
    \begin{tabular}{clllll}
\toprule
    \textbf{Duration} & \textbf{Methods} & \textbf{NSS}~$\uparrow$ & \textbf{CC}~$\uparrow$ & \textbf{KL}~$\downarrow$ & \textbf{SIM}~$\uparrow$ \\
\midrule
    \multirow{5}{*}{3\,s} & DCSM~\cite{bao2020human} & 0.678    & 0.293    & 1.228    & 0.409 \\
    & MD-SEM~\cite{fosco2020much} & 1.086 & 0.474 & \underline{0.840} & \underline{0.485} \\
    & DVS~\cite{matzen2017data} & 1.106   & 0.456    & 0.933    & 0.449 \\
    & MASSVIS-MDEFD & \underline{1.208} & \underline{0.502} & 1.250 & 0.476 \\
    & MD-EAM~(Ours) & \textbf{1.406$^{***}$} & \textbf{0.586$^{***}$} & \textbf{0.754$^{***}$} & \textbf{0.516$^{***}$} \\\midrule
    \multirow{5}{*}{5\,s} & DCSM~\cite{bao2020human} & 0.721  & 0.371 & 0.900 & 0.492 \\
    & MD-SEM~\cite{fosco2020much} & 0.908 & 0.479 & 0.709 & 0.527 \\
    & DVS~\cite{matzen2017data} & \textbf{1.031}    & \underline{0.510}   & \textbf{0.681}    & \textbf{0.531} \\
    & MASSVIS-MDEFD & 0.932 & 0.448 & 1.119 & 0.491 \\
    & MD-EAM~(Ours) & \underline{1.024} & \textbf{0.514} & \underline{0.689} & \underline{0.530}  \\
\bottomrule
    \end{tabular}
    \label{table:saliency}
\end{table*}

\textit{Metrics.}
We use four popular metrics for evaluating performance: Normalized Scanpath Saliency~(NSS), Pearson’s Correlation Coefficient~(CC), Kullback-Leibler divergence~(KL), and Similarity or histogram intersection~(SIM). NSS is calculated on fixation maps, while CC, KL and SIM are calculated on saliency maps.

\textit{Results.}
Table~\ref{table:saliency} demonstrates the performance of saliency prediction methods using ground-truth duration of 3\,s and 5\,s. 
Our method ranks first in all metrics in 3\,s duration, and is tied with DVS~\cite{matzen2017data} under 5\,s duration.

\subsection{Ablation Studies}
We further carried out two ablation studies to evaluate the effectiveness of our model. First, we replaced our \methodSaliencyNameShort with several saliency methods to see the influence of the saliency model on scanpaths. Then, we remove components in our scanpath sampling strategy to analyse how each component contributes to the final model.

\begin{table*}[!t]
    \centering
    \caption{Ablation study on saliency encoder and sampling strategy. All methods are evaluated with 5-second ground truth in terms of Sequence Score~(SS), ScanMatch~(SM), Dynamic Time Warping~(DTW), and scaled Time Dimension Embedding~(sTDE). Best results are shown in \textbf{bold}, best baselines are \underline{underlined}. Stars indicate statistical significance of the difference between our Full Model and the best baseline model (**: p\,<\,.01; ***: p\,<\,.001).}
    \begin{tabular}{lllll}
\toprule
    \textbf{Methods} & \textbf{SS}~$\uparrow$ & \textbf{SM}~$\uparrow$ & \textbf{DTW~(2D)}~$\downarrow$ & \textbf{sTDE}~$\uparrow$ \\
\midrule
    Saltinet~\cite{assens2017saltinet} & 0.388 & 0.331 & 12758.51  & 0.875 \\
    DVS~\cite{matzen2017data}~+~Saltinet & 0.398 & \underline{0.381} & 7762.77 & 0.881 \\
    MD-SEM~\cite{fosco2020much}~+~Saltinet & 0.396 & 0.325 & 7932.85  & 0.897 \\ %
    MD-EAM~+~Saltinet & \underline{0.436} & 0.330 & \underline{7286.56} & \underline{0.903} \\
\midrule
    w/o Slice Allocation & 0.437 & 0.332 & 7213.87 & 0.903 \\
    w/o Duration Prediction & 0.445 & 0.367 & 6570.35 &  0.905 \\
    Full Model & \textbf{0.446$^{***}$} & \textbf{0.387$^{***}$} & \textbf{6529.11$^{***}$} & \textbf{0.906} \\
\bottomrule
    \end{tabular}
    \label{table:abl_saliency}
\end{table*}

\subsubsection{Saliency Model}
We compared the performance of our \methodSaliencyNameShort with two saliency models, that is, DVS~\cite{matzen2017data} and MD-SEM~\cite{fosco2020much}, by plugging in the post-processing algorithm of Saltinet to each of the saliency models.
Table~\ref{table:abl_saliency} shows the effectiveness of our \methodSaliencyNameShort, which outperforms all the other methods in all scanpath metrics.

\subsubsection{Scanpath Sampling Strategy}
We evaluated the scanpath sampling strategy by removing its components. 
We replaced the fixation assigning strategy by evenly sampling fixations for each slice of multi-duration attention maps~\cite{assens2017saltinet}, and removed our Slice Allocation strategy. Table~\ref{table:abl_saliency} shows that all metrics improved by adding Slice Allocation to the full model.

\subsection{User Study}
To gain further insights, we designed a study in which participants had to qualitatively compare human, ground-truth scanpaths from the MASSVIS dataset to predictions from Saltinet, DCSM, PathGAN, and UMSS (ours).
Additionally, we included a second, randomly selected ground-truth scanpath as a Human baseline.
For each trial in the evaluation, we randomly selected one human scanpath from the same visualisation as the \textit{Target}.
We randomly sampled scanpaths for the three baselines where multiple scanpaths are existed (Saltinet, UMSS, and Human), while PathGAN and DSCM produced only a single scanpath. 
Study participants were asked to compare the five baselines to the human \textit{Target} by ranking the generated scanpaths from 1 to 5, where 1\,=\,most similar and 5\,=\,most dissimilar (see Figure~10 in supplementary material). 
The presentation order of the five baselines was counterbalanced using a latin-square study design.
The study contained 40 trials, i.e. 40 visualisations from the MASSVIS evaluation set.
The duration of the entire study was around 30\,minutes and participants were compensated €\,10 for participation. 

We recruited ten researchers from our university who were familiar with gaze data and had experience in eye-tracking studies.
Results showed that the Human baseline had the highest mean ranking of 1.53 ($\sigma$\,=\,0.81). 
The second mean ranking was achieved by UMSS (ours) with 1.98 ($\sigma$\,=\,0.96).
Saltinet, DCSM, and PathGAN had a mean ranking of 3.58\,($\sigma$\,=\,0.97), 3.73\,($\sigma$\,=\,1.01), and 4.18\,($\sigma$\,=\,1.01).
The highest mean expert rating of the three scanpath prediction baselines is only 3.66~($\sigma$\,=\,0.99), which is significantly lower than UMSS~(t\,(638)\,=\,21.89, p\,<\,0.001).
Five examples from our study are illustrated in \autoref{fig:qualitative}.
Upon completing all trials, we asked participants to provide qualitative feedback on the most important characteristics they used in their subjective evaluation.
From the subjective feedback that justified similarity of scanpaths, participants often mentioned ``Text labels in the Visualization'', ``the movement of the path and the area it covered''.
Some frequently mentioned characteristics that made scanpaths dissimilar were ``Too crowded scan paths, too widespread scanpaths'' and ``Frequent and fast changes in direction''.

\section{Discussion}\label{sec:discussion}

\textit{Experiment Results.}
To the best of our knowledge, our method is the first to predict human scanpaths on information visualisations.
We first compared \methodNameShort to three state-of-the-art methods (PathGAN~\cite{assens2018pathgan}, DCSM~\cite{bao2020human}, and Saltinet~\cite{assens2017saltinet}) using five popular evaluation metrics: the Sequence Score~\cite{yang2020predicting}, DTW~\cite{muller2007dynamic}, sTDE~\cite{zanca2018fixatons, wang2011simulating}, ScanMatch~\cite{cristino2010scanmatch}, and MultiMatch~\cite{jarodzka2010vector} (see \autoref{table:quantitative}). 
In terms of the Sequence Score, which converts fixations to characters that represent semantic regions, our method outperformed the others with a relative improvement of 11.5\,\% by \textit{mean} and 10.33\,\% by \textit{best}.
Our method also achieved the best performance for ScanMatch and for two dimensions of MultiMatch~(direction and position).
As for the remaining evaluation metrics, our method generally ranked second place, and there was no single method that outperformed all others for all metrics.
For predicting fixation durations, our method ranks second.
To our surprise, DCSM~\cite{bao2020human} exceeds the human baseline (0.755 vs 0.730), which indicates that the variance of fixation duration across human viewers is rather large.
However, it is important to note that current scanpath evaluation metrics have been developed for natural scenes.
Therefore, it is possible that some metrics do not work as well for quantifying scanpath quality on information visualisations.
This naturally leads to the question of \emph{Which method is better on information visualisations?}, and more fundamentally, \emph{Which evaluation metrics are suited for scanpath prediction on information visualisations?}

\textit{Scanpath Metrics.}
As discussed above, current scanpath metrics are devised for natural scenes and have not been tested on information visualisations. Moreover, the quantitative rankings (\autoref{table:quantitative}) and human ratings (\autoref{fig:qualitative}) disagree with which method can produce human-like scanpaths better. Therefore, it is necessary to take a deep look into how well current scanpath metrics work on information visualisations.
Our user study gave a clear answer to which method predicts scanpaths that are perceived as most natural or human-like, and which metrics are closer to human ratings on information visualisations.
Our method is the second most comparable ($\mu$\,=\,1.98, $\sigma$\,=\,0.96), directly following the human baseline ($\mu$\,=\,1.54, $\sigma$\,=\,0.80), and is significantly closer to human scanpaths than any existing scanpath prediction baselines.
The scanpaths predicted by \methodNameShort are visually more similar to human scanpaths, which is in agreement with expert ratings from our user study.
Saltinet~\cite{assens2017saltinet} is the next preferred method, but closer visual inspection of the scanpaths reveals that many fixations are scattered throughout the image, including also in white spaces~(see \autoref{fig:qualitative}).
The scanpaths produced by DCSM~\cite{bao2020human} that achieved the highest score in terms of DTW, as well as PathGAN~\cite{assens2018pathgan} that achieved the highest score for sDTE and two dimensions of MultiMatch, are very different from a qualitative point of view: 
Fixations predicted by DCSM are clustered in several smaller regions, while those predicted by PathGAN are clustered in the centre of the visualisation~(see \autoref{fig:qualitative}). 
This shows that DCSM and PathGAN fail to predict scanpaths that are rated as human-like, although they rank first in some scanpath metrics.

After comparing the quantitative results and our user study (see \autoref{fig:qualitative} and Figure~\,6 in supplementary material), we noticed that the sTDE, DTW, and MultiMatch metrics are often in contradiction with the expert ratings from our user study. These metrics can achieve the highest scores even if expert ratings are low.
This phenomenon explains why our method achieved promising results in ScanMatch and Sequence Score, but didn't outperform the other methods in sTDE, DTW, and MultiMatch (see \autoref{table:quantitative}).
Taking these quantitative and qualitative findings together, only a few of existing metrics (Sequence Score and ScanMatch) agree with expert ratings when evaluating predicted scanpaths on information visualisations. 
Metrics that evaluate pixel-wise distances between scanpaths, such as MultiMatch, DTW and sTDE, do not fully capture the quality of human scanpaths. 
This is, in part, due to the nature of the visual stimuli.
For natural images, information is often less structured and fixations can be found anywhere.
In contrast to natural images, the semantic regions in information visualisations are separated by the white spaces, and fixations are much more likely to be inside these semantic regions, rather than white spaces. 
In contrast, metrics that take the semantic regions of fixations into account, such as the Sequence Score, are more consistent with expert ratings.
The auspicious results of our user study suggest that~--~despite the fact that some existing metrics seem to show that our method does not outperform others~--~\methodNameShort is a significant step towards predicting scanpaths on information visualisations that are more natural and human-like.

\textit{Scanpaths and Saliency.}
\autoref{table:saliency} shows that \methodSaliencyNameShort achieves the highest saliency metrics for the 3-second ground truth, and shares the first place with DVS~\cite{matzen2017data} for the 5-second ground truth.
Multi-duration saliency methods have an advantage in flexibility, that is, \methodSaliencyNameShort is competitive for every duration.
Furthermore, \autoref{table:abl_saliency} shows that \methodSaliencyNameShort outperforms the remaining Saltinet-based methods in Sequence Score, DTW, and sTDE.
This indicates that for those methods that sampled from saliency maps, the better the saliency maps, the better scanpaths can be predicted.
In summary, this work predicts human-like scanpaths on information visualisations and shows powerful performance in multi-duration saliency prediction. 

\textit{Gaze Behaviour on Information Visualisations.}
In Section 3, we analysed gaze behaviour on the MASSVIS dataset and concluded that viewers tend to focus on a visualisation differently depending on how long they have been observing it.
We found that the Sequence Scores across viewers was only 0.4\,--\,0.6 (see \autoref{fig:analysis_fixationdensity}).
This suggests that viewers' gaze behaviour on information visualisations is subject to a considerable amount of variability.
Another finding specific to information visualisations is that different visualisation elements are salient under different viewing durations.
This explains why our method reaches state-of-the-art performance. 
\methodSaliencyNameShort learns the dynamics of gaze behaviour on information visualisations, and minimises the information loss when generating scanpaths from the saliency maps.

\subsection{Limitations}
Due to the data scarcity problem of gaze data under free-viewing condition on information visualisations,
we only analysed and trained our scanpath prediction model for memorability tasks. Since viewers were asked to memorise as much information as possible, attention towards textual regions such as titles might be preferable than free-viewing conditions. 
Given that top-down attention plays an important role in visualisations, it is crucial to understand top-down attention behaviours.

We also assumed that all elements in information visualisations are known as prior knowledge. This is a reasonable assumption on visualisations, since they are artificial and contain well-structured data. However, incorrect annotations or detection of its constituting elements will decrease the performance of our scanpath sampling strategy. 
Element information from MASSVIS is manually annotated, but, in practice, a good object detection model to automatically retrieve annotations is necessary to visually parse and decode information visualisation that do no have these annotations.

\section{Conclusion}

In this work, we proposed \methodNameLong, the first method designed to predict realistic scanpaths on information visualisations. We systematically analysed the element-level attention dynamics on information visualisations, and revealed consistencies across visualisations and viewers. Our novel multi-duration element attention maps and data-driven sampling strategy allowed our model to generate scanpaths of significantly better quality than previous methods. 
Our method reached the state of the art on both saliency and scanpath prediction tasks on MASSVIS.
In conclusion, our work provided a new perspective towards scanpath prediction on information visualisations and points towards novel computational methods to better predict human scanpaths without the need for eye tracking equipment.

\ifCLASSOPTIONcompsoc
  \section*{Acknowledgments}
\else
  \section*{Acknowledgment}
\fi

Y. Wang was funded by the Deutsche Forschungsgemeinschaft~(DFG, German Research Foundation)~-~Project-ID 251654672~-~TRR~161.
M. B\^{a}ce was funded by a Swiss National Science Foundation (SNSF) Early Postdoc. Mobility Fellowship (grant number 199991).
A. Bulling was funded by the European Research Council (ERC; grant agreement 801708).

The authors would like to thank Anelise Newman for providing training and evaluation details about MDSEM, Wentao Bao for providing DCSM predictions on MASSVIS, Dominike Thomas for paper editing support, Sruthi Radhakrishnan and Saiteja Malyala for developing a scanpath visualisation toolbox for MASSVIS, Lei Shi for assistance in the review process, as well as Zhiming Hu and Nils Rodrigues for helpful comments on an earlier paper draft.

\ifCLASSOPTIONcaptionsoff
  \newpage
\fi

\bibliographystyle{IEEEtranN}
\bibliography{main}

\begin{IEEEbiography}[{\includegraphics[width=1in,height=1.25in,clip,keepaspectratio]{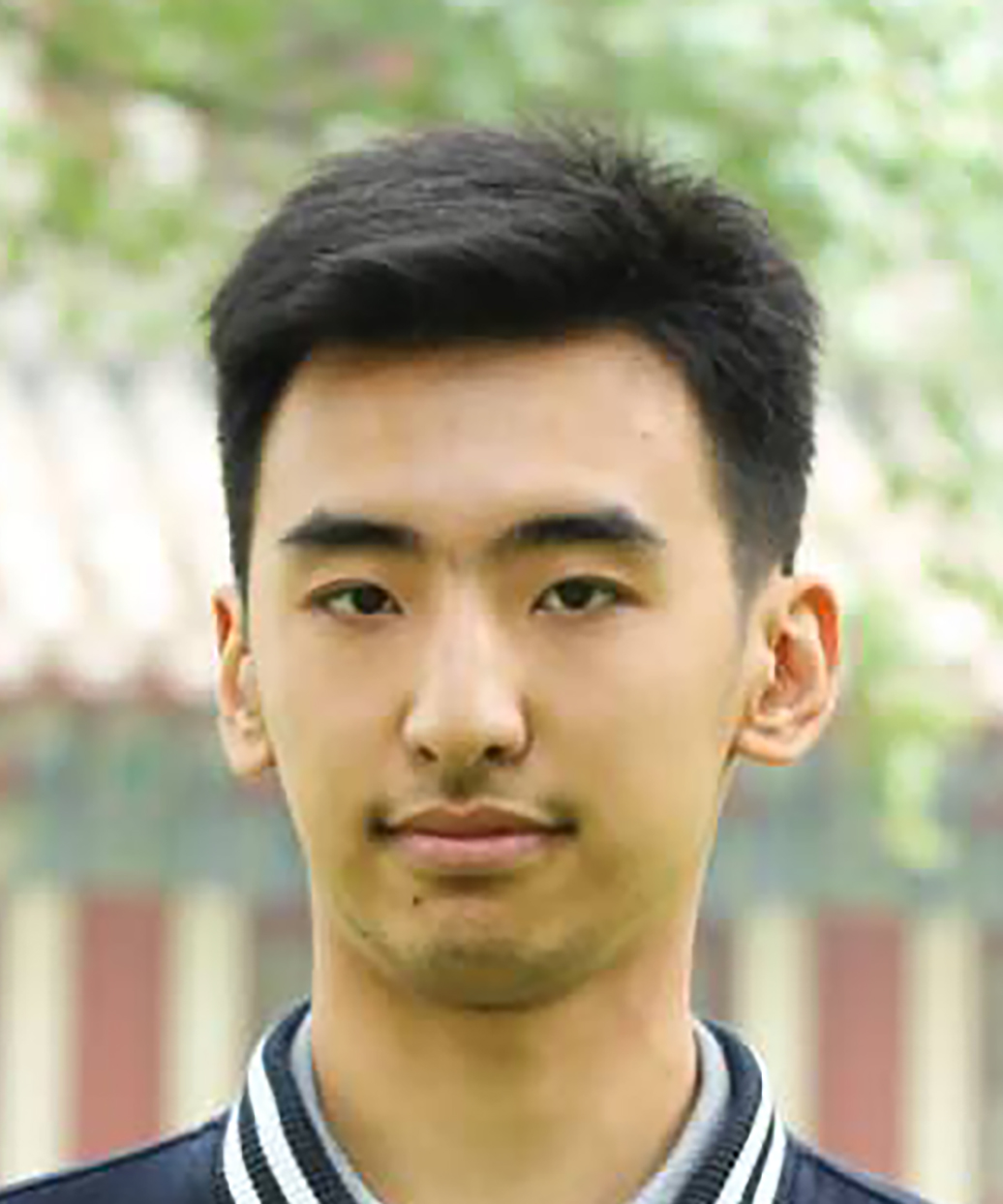}}]{Yao Wang}
is a PhD student at the University of Stuttgart, Germany. He received the BSc degree in Intelligence Science and Technology and MSc degree in Computer Software and Theory both from Peking University, China, in 2017 and 2020, respectively. His research interests include computer vision and human-computer interaction, with a focus on visual attention modelling on information visualizations.
\end{IEEEbiography}

\begin{IEEEbiography}[{\includegraphics[width=1in,height=1.25in,clip,keepaspectratio]{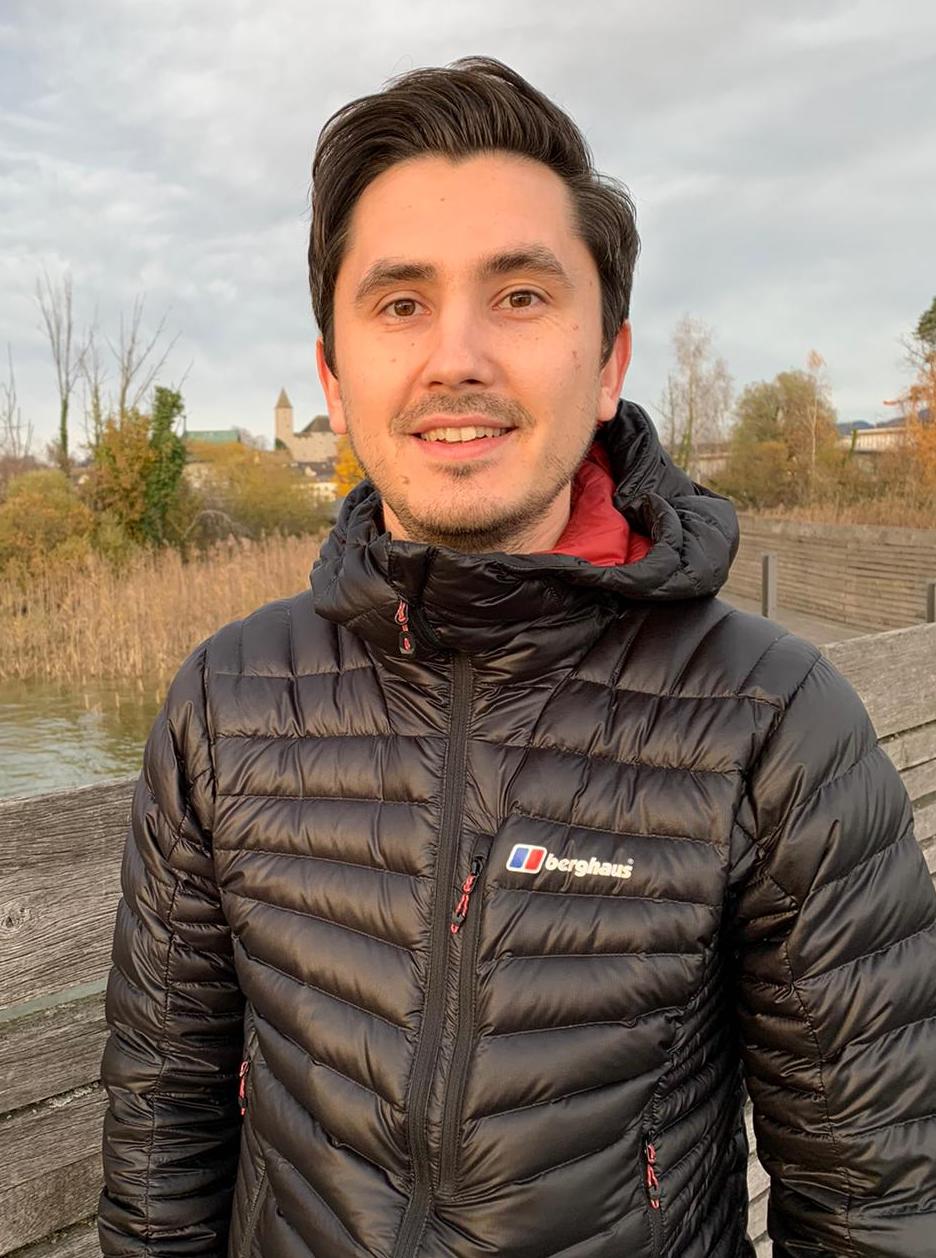}}]{Mihai~B\^{a}ce} is a post-doctoral researcher in the Perceptual User Interfaces group at the University of Stuttgart, Germany. He did the PhD at ETH Zurich, Switzerland, at the Institute for Intelligent Interactive Systems. He received the MSc degree in Computer Science from École Polytechnique Fédérale de Lausanne, Switzerland, and the BSc degree in Computer Science from the Technical University of Cluj-Napoca, Romania. His research interests include computational Human-Computer Interaction with a focus on sensing and modelling user attention.
\end{IEEEbiography}

\begin{IEEEbiography}[{\includegraphics[width=1in,height=1.25in,clip,keepaspectratio]{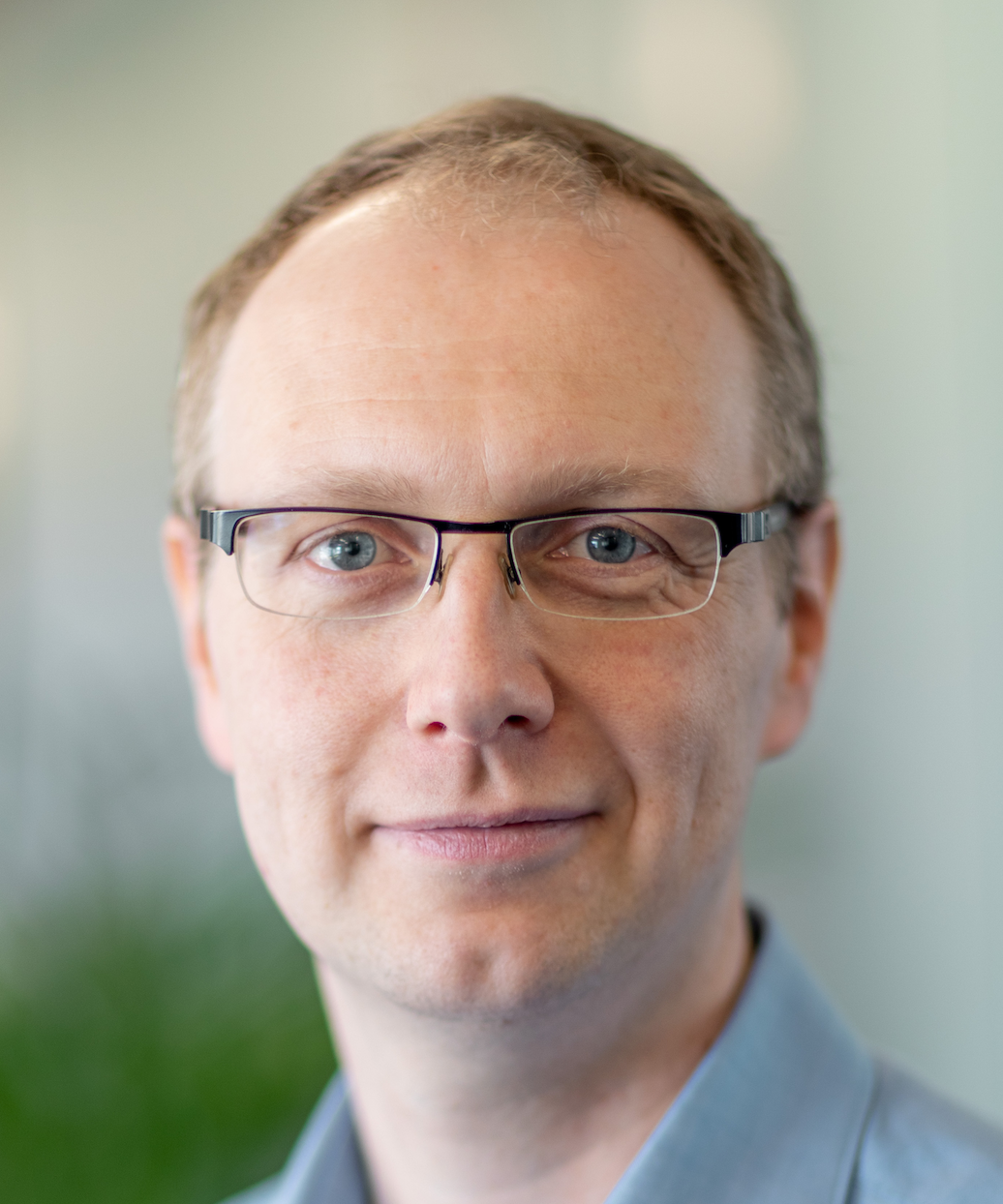}}]{Andreas~Bulling}
is Full Professor of Computer Science at the University of Stuttgart, Germany, where he directs the research group "Human-Computer Interaction and Cognitive Systems". He received the MSc degree in Computer Science from the Karlsruhe Institute of Technology, Germany, in 2006 and the PhD degree in Information Technology and Electrical Engineering from ETH Zurich, Switzerland, in 2010. Before, Andreas Bulling was a Feodor Lynen and Marie Curie Research Fellow at the University of Cambridge, UK, and a Senior Researcher at the Max Planck Institute for Informatics, Germany. His research interests include computer vision, machine learning, and human-computer interaction.
\end{IEEEbiography}

\vfill

\enlargethispage{-5in}

\end{document}


\title{Supplementary Material for ``Scanpath \\ Prediction on Information Visualisations''}

\author{Yao~Wang,~Mihai~B\^{a}ce,~and~Andreas~Bulling%
}

\markboth{IEEE Transactions on Vizualisation and Computer Graphics}%
{Wang \MakeLowercase{\textit{et al.}}: Scanpath Prediction on Information Visualisations}

\maketitle

\IEEEdisplaynontitleabstractindextext

\IEEEpeerreviewmaketitle

\ifCLASSOPTIONcaptionsoff
  \newpage
\fi

\begin{figure}[ht]
  \centering
    \includegraphics[width=\linewidth]{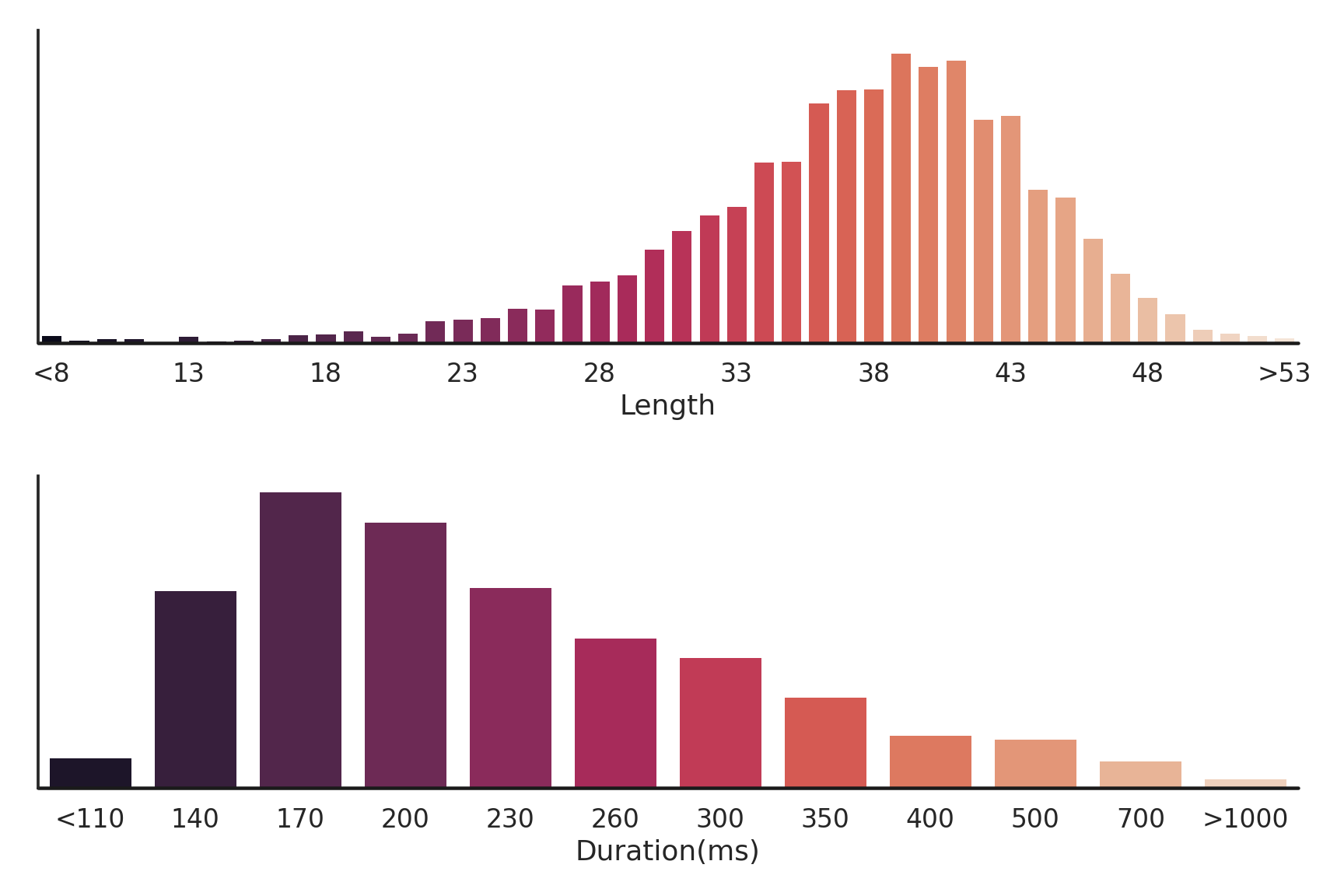}
  \caption{Distributions of scanpath length (top) and fixation duration (bottom) from the MASSVIS~\cite{borkin2013makes,borkin2015beyond} dataset.}
  \label{fig:LenDurDistribution}
\end{figure}

\begin{figure}[ht]
  \subfloat[]{\includegraphics[width=0.5\linewidth]{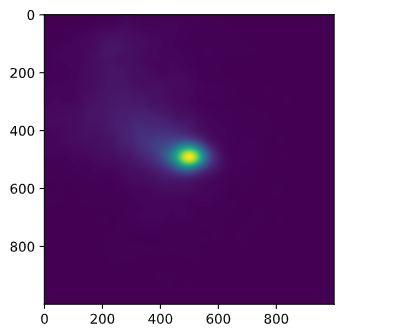}}
  \subfloat[]{\includegraphics[width=0.5\linewidth]{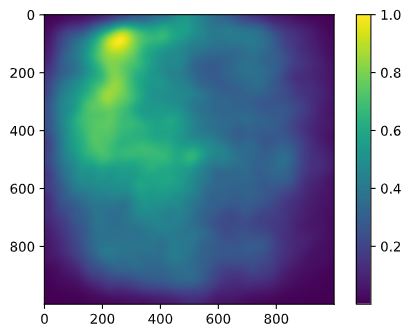}}
  \caption{Accumulated fixation distribution from the MASSVIS dataset. (a) The first fixations of all viewers. (b) The rest fixations except the first fixations of all viewers. There is a strong centre bias within the first fixations across all viewers. This is due to the experiment setting, where a fixation cross shows up right before the image appears on the screen~\cite{borkin2015beyond}.
  }
  \label{fig:analysis_centerbias}
\end{figure}

\begin{figure}[t]
  \centering
    \includegraphics[width=\linewidth]{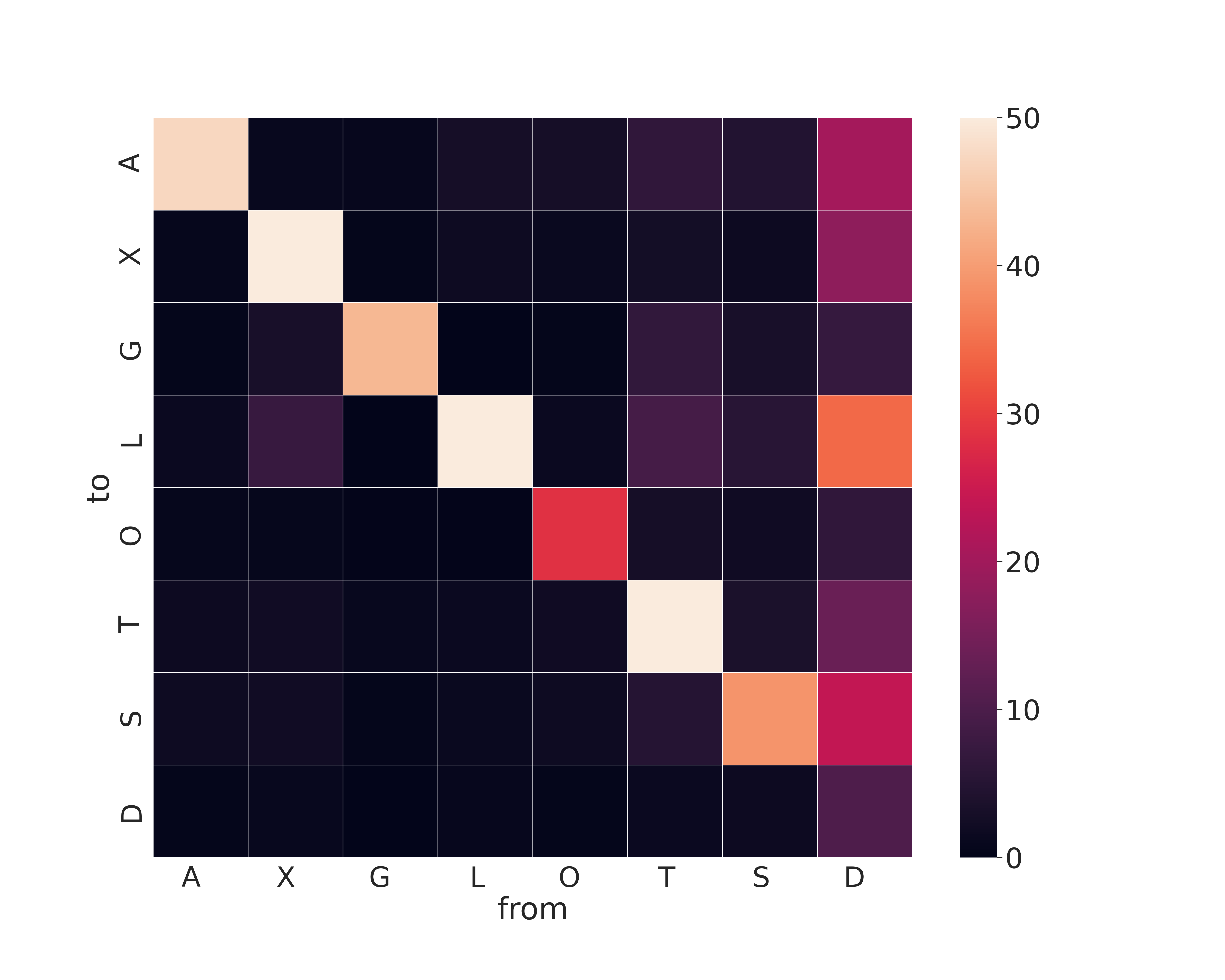}
    \includegraphics[width=\linewidth]{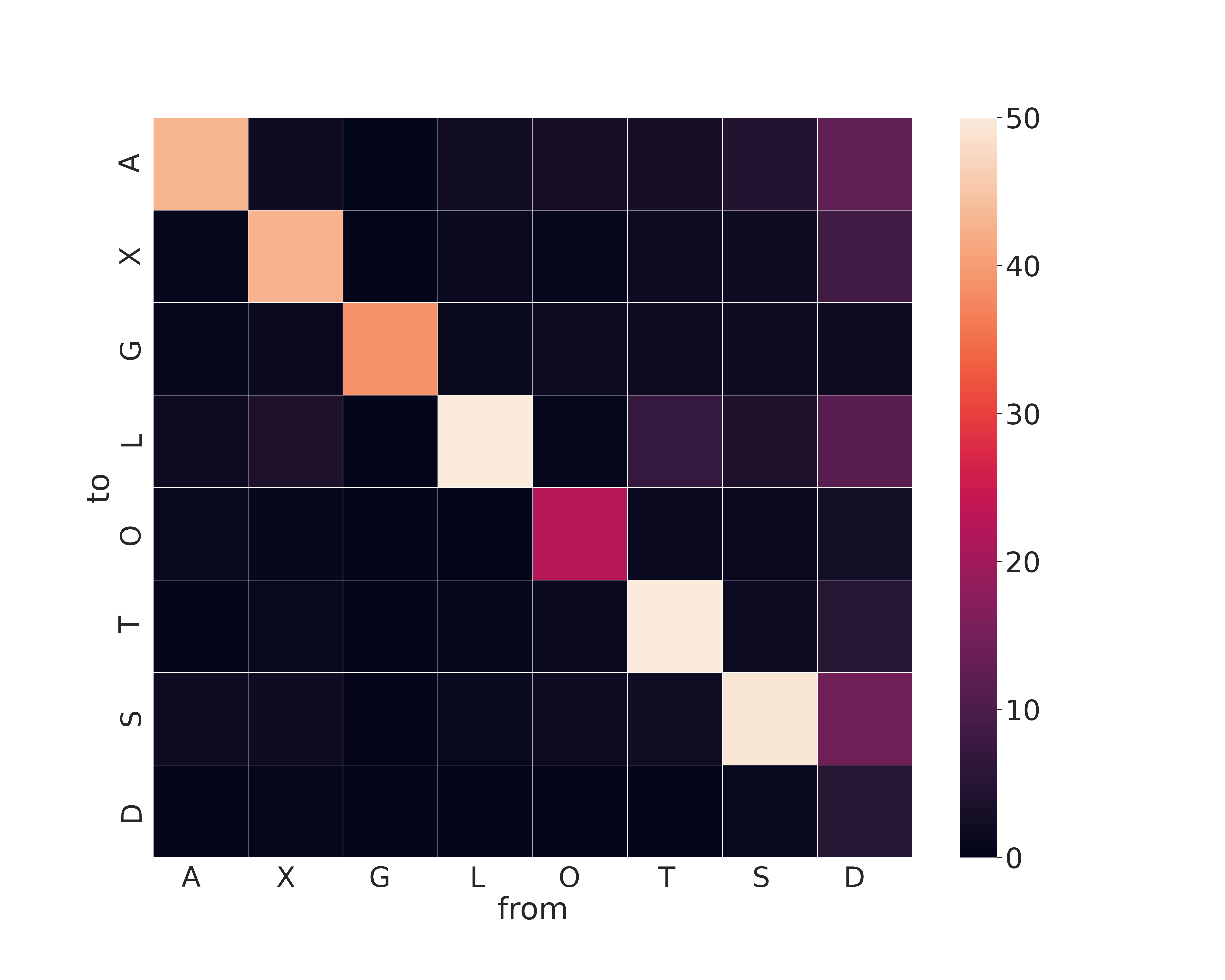}
  \caption{Transition matrices of two viewers in MASSVIS. Viewers tend to look at Title and Legend continuously before jumping to other regions, while they tend to read Data in cooperation with Annotation, Axis, and Legend. A: Annotation, X: Axis, G: Graphics, L: Legend, O: Object, T: Title, S: Source etc., D: Data.}
  \label{fig:transition_matrices}
\end{figure}

\begin{figure*}[t]
  \centering
    \includegraphics[width=0.4\linewidth]{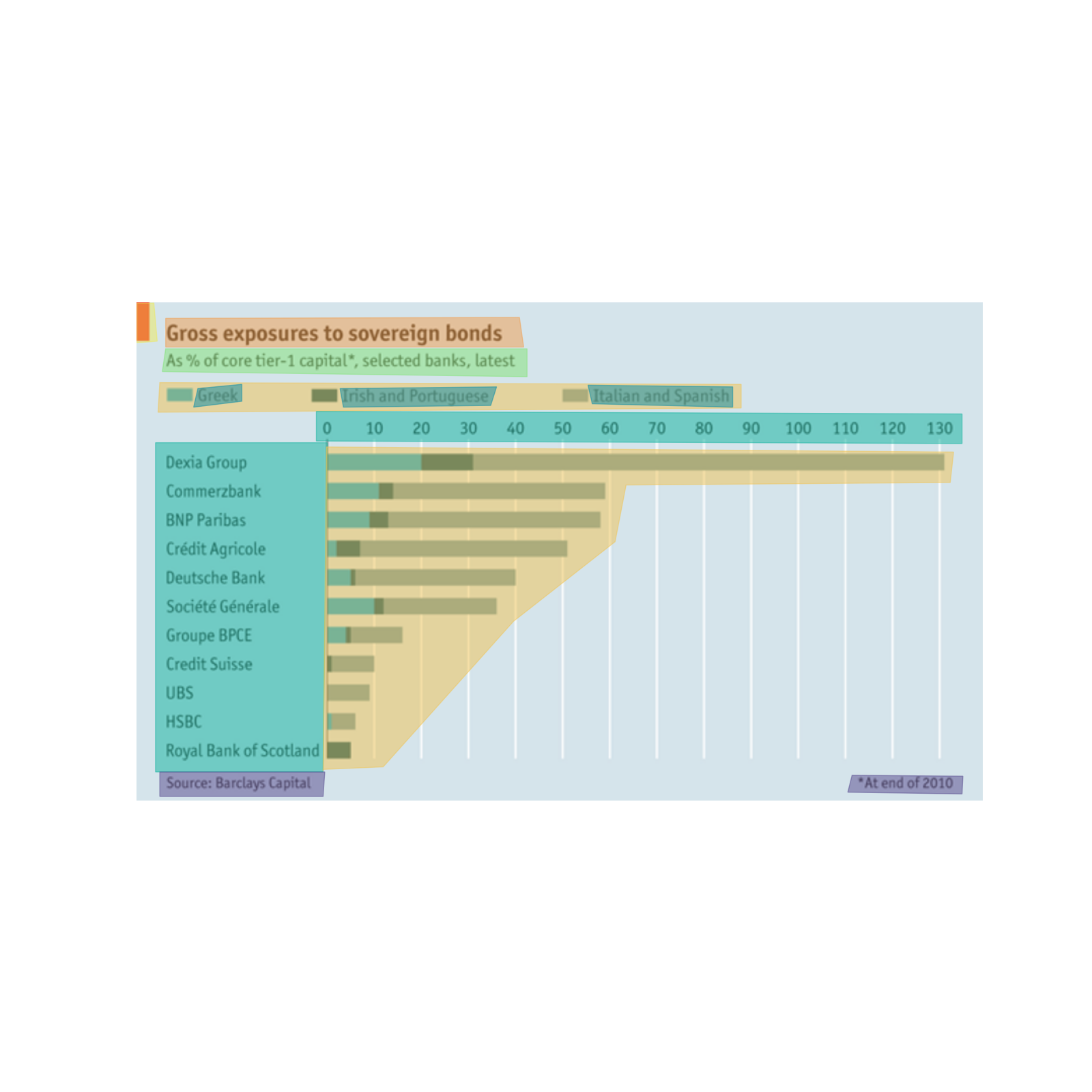}
    \includegraphics[width=0.4\linewidth]{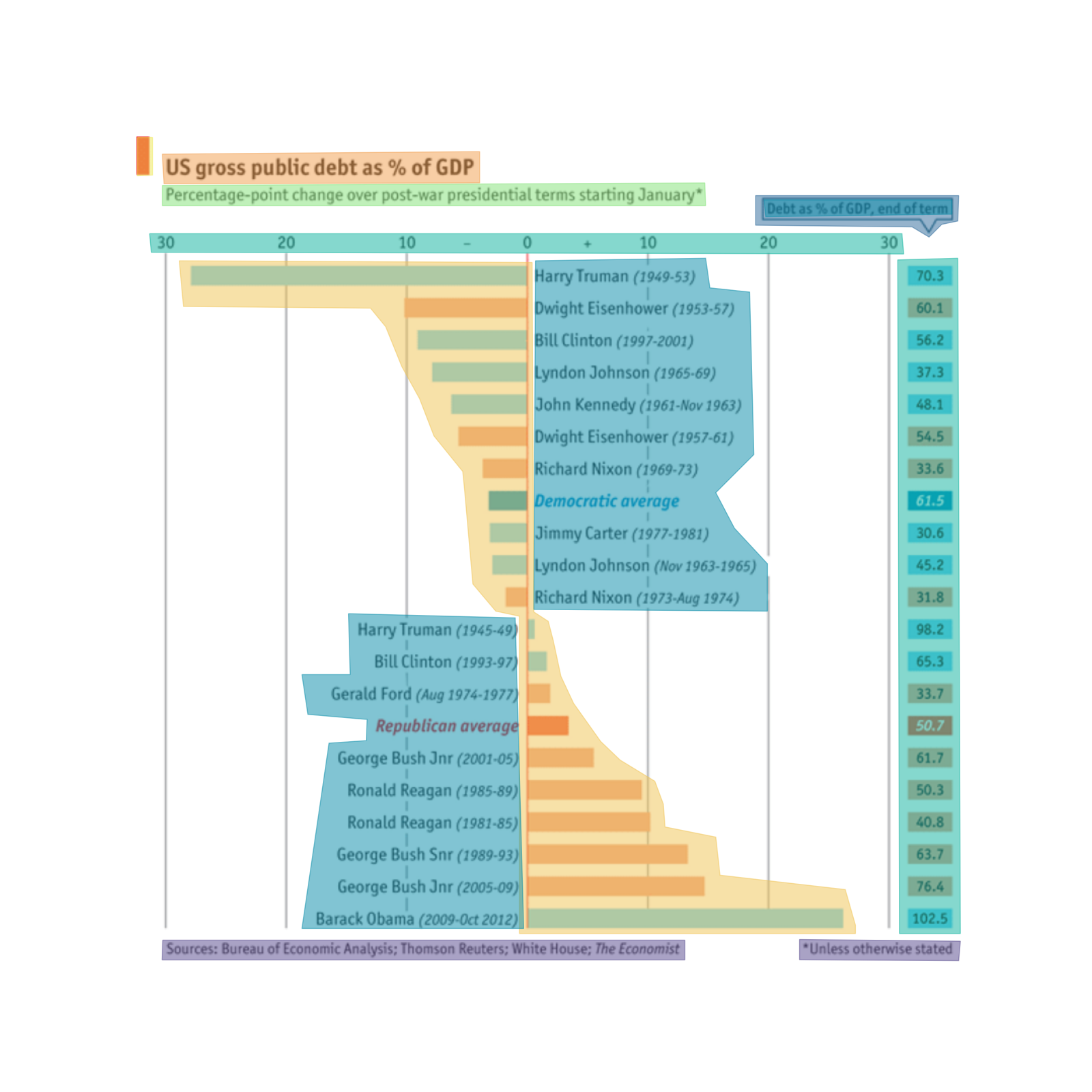}
    \includegraphics[width=0.4\linewidth]{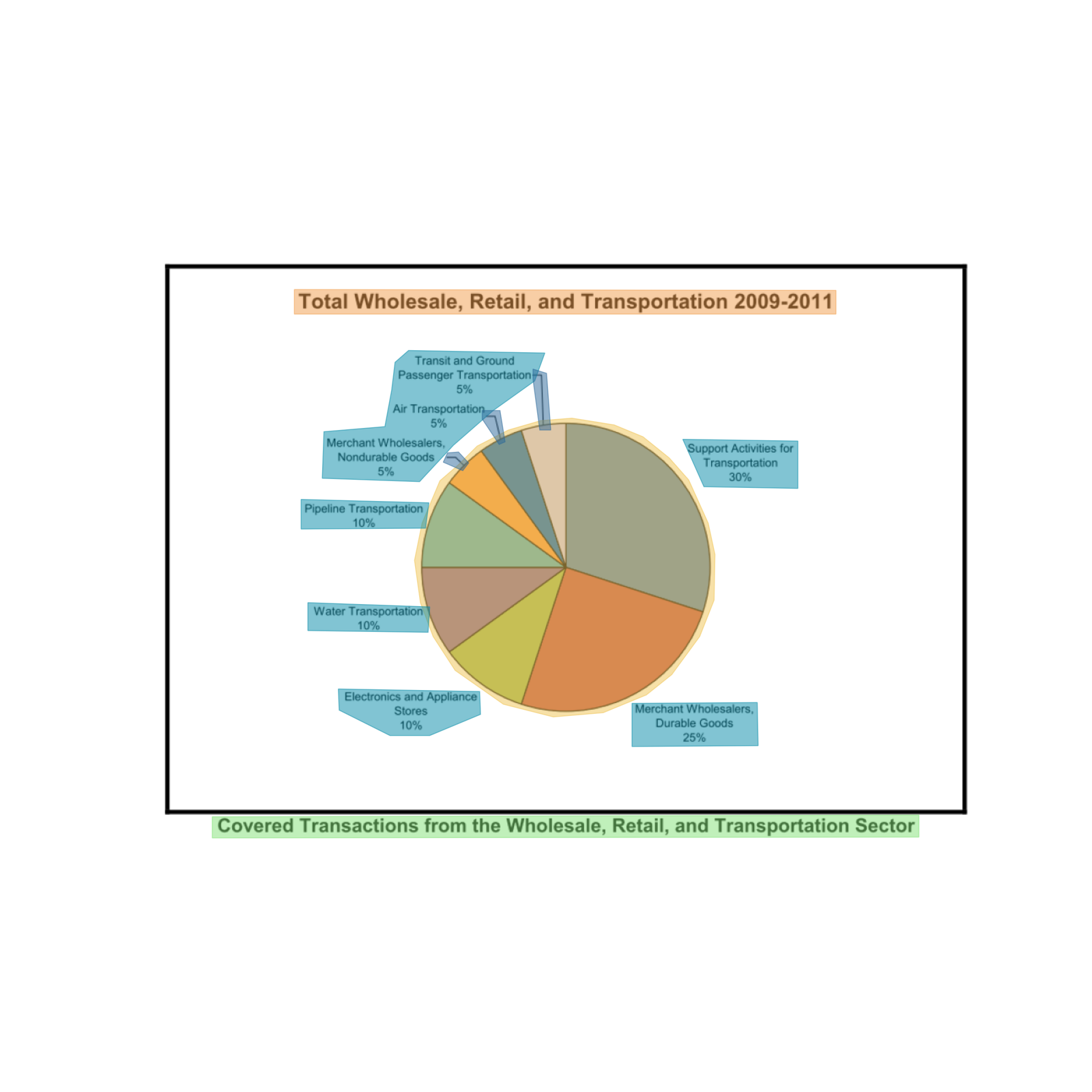}
    \includegraphics[width=0.4\linewidth]{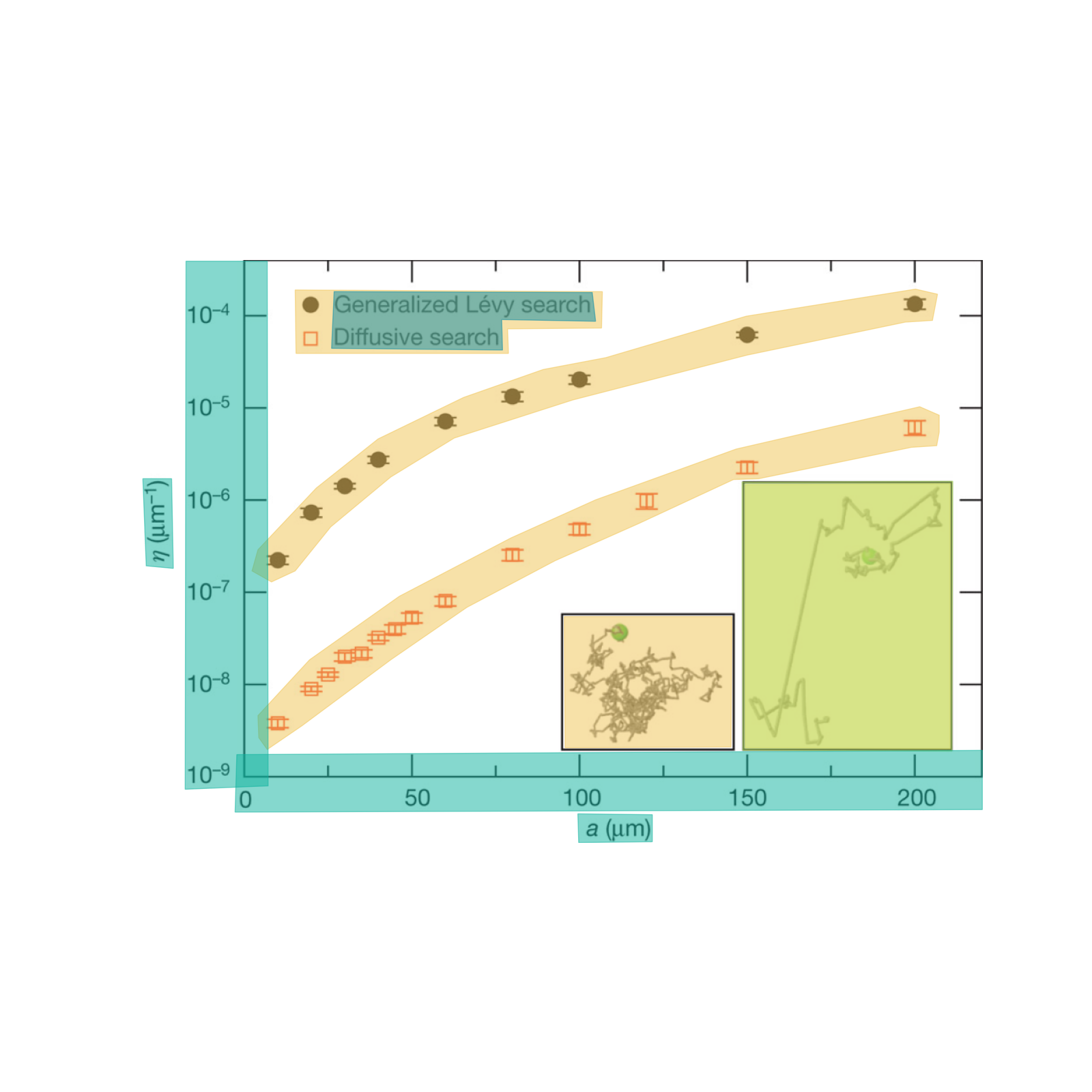}
    \includegraphics[width=0.4\linewidth]{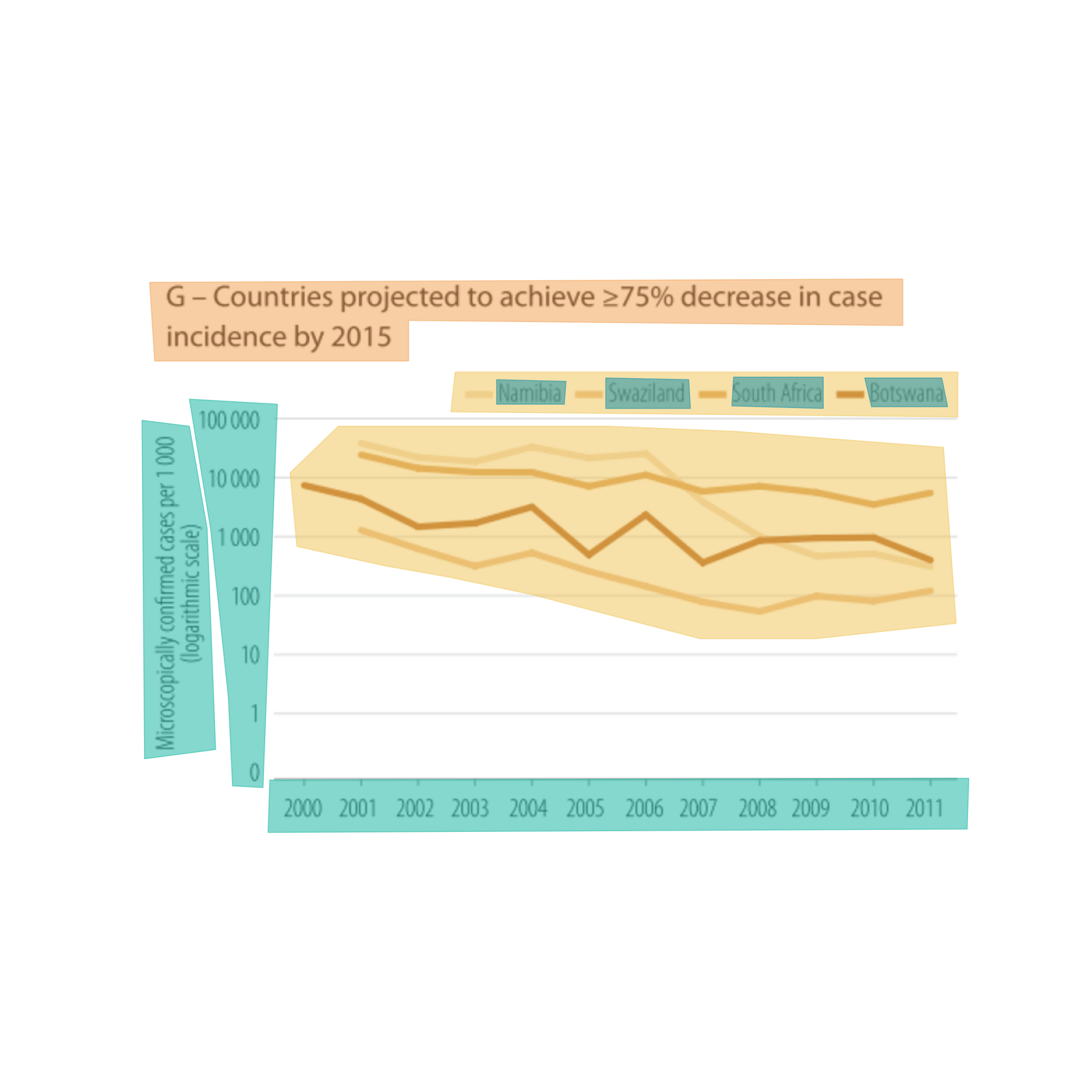}
    \includegraphics[width=0.4\linewidth]{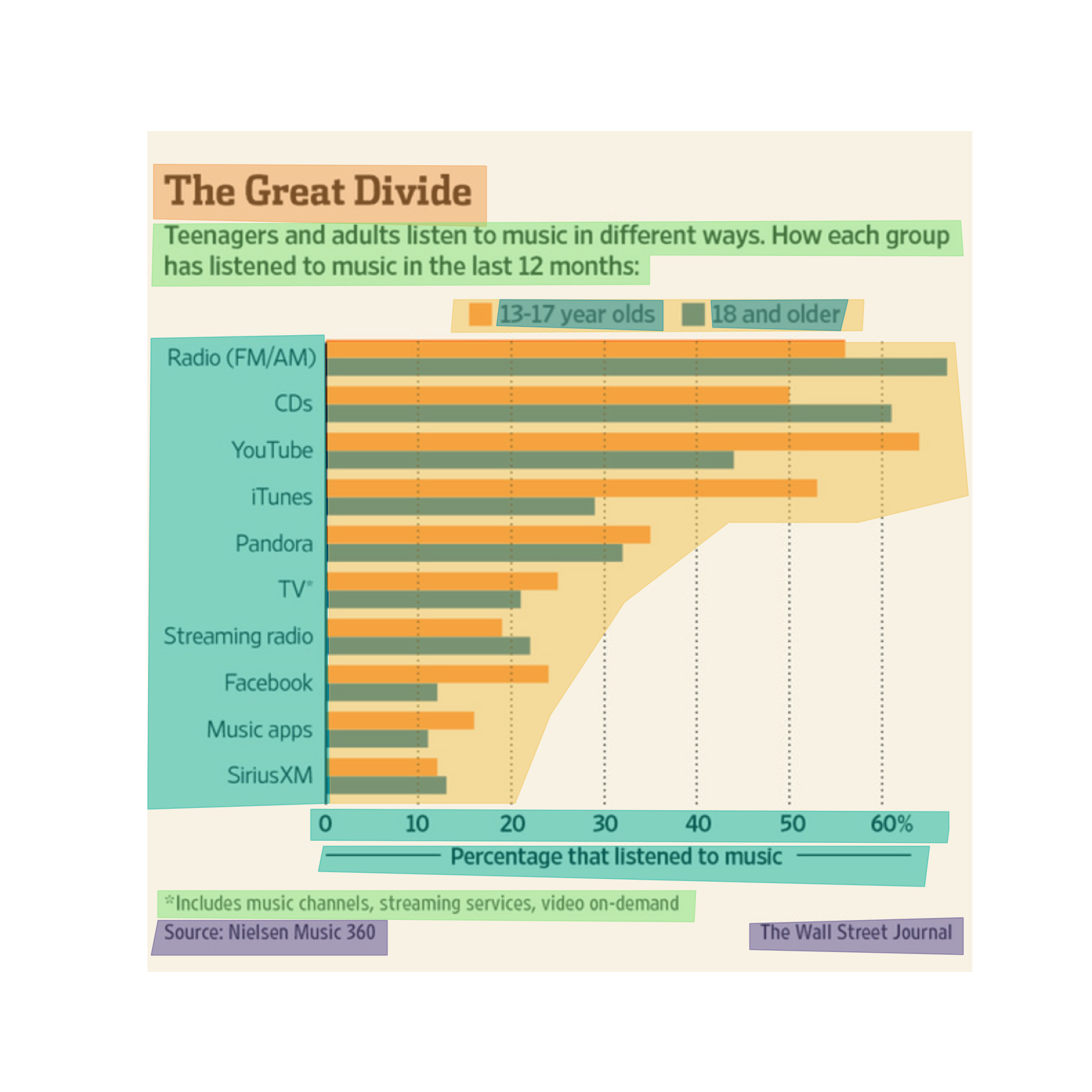}
  \caption{Example visualisations from the MASSVIS dataset as well as visualisation element annotations highlighted in colour. Each visualisation element (e.g. Title or Label) have a unique colour and the colouring policy is consistent with Figure~2 from the main manuscript.}
  \label{fig:example_annotations}
\end{figure*}

\begin{figure*}[ht]
  \centering
    \subfloat[Example stimuli]{\includegraphics[width=0.66\linewidth]{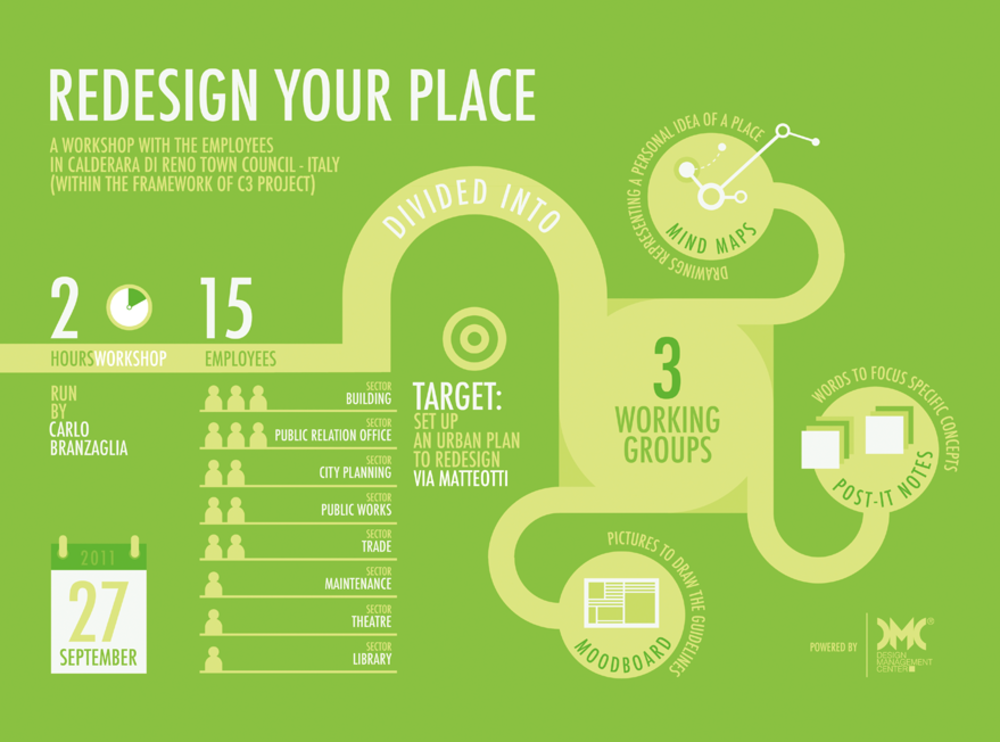}}\\
    \subfloat[Saliency map of 0.5\,s]{\includegraphics[width=0.33\linewidth]{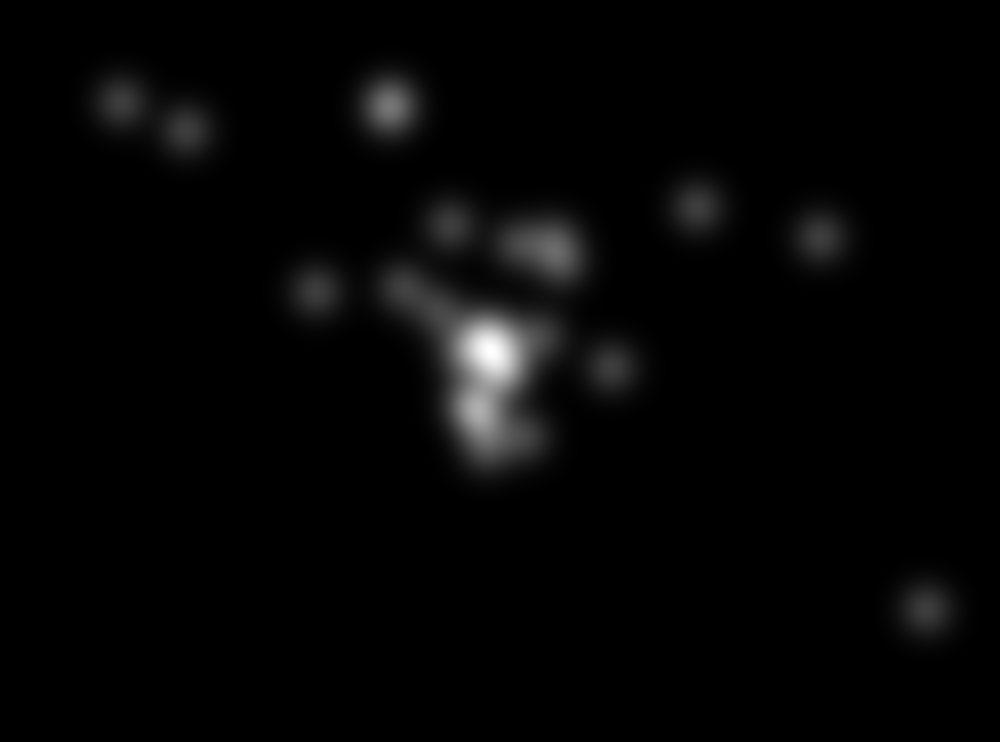}}
    \subfloat[EFD map of 3\,s]{\includegraphics[width=0.33\linewidth]{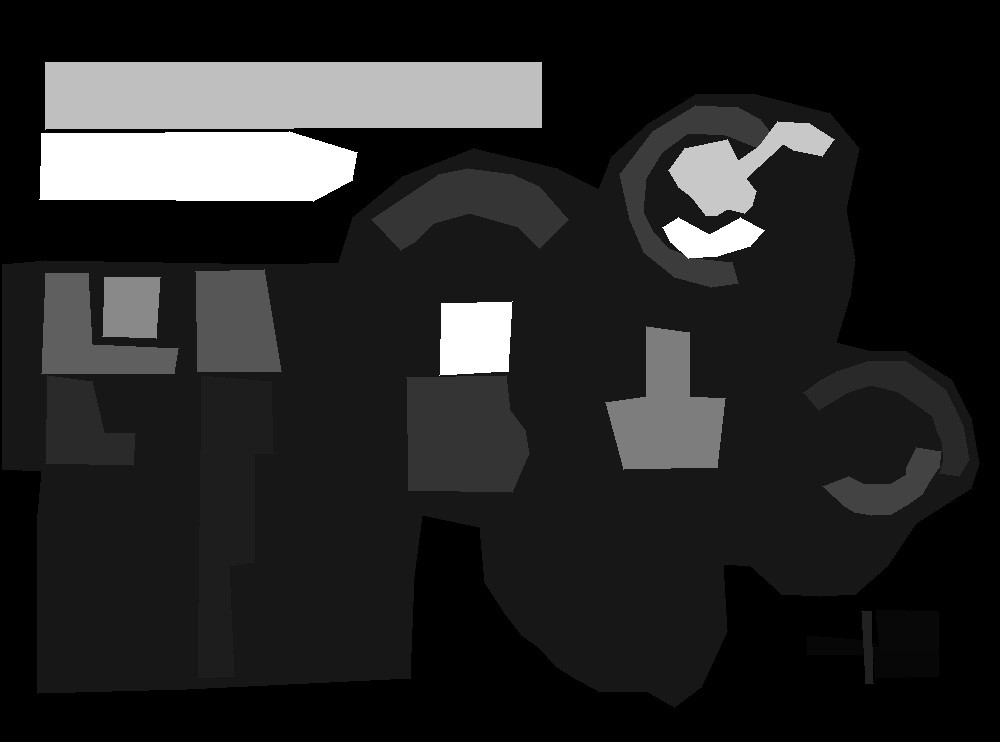}}
    \subfloat[EFD map of 5\,s]{\includegraphics[width=0.33\linewidth]{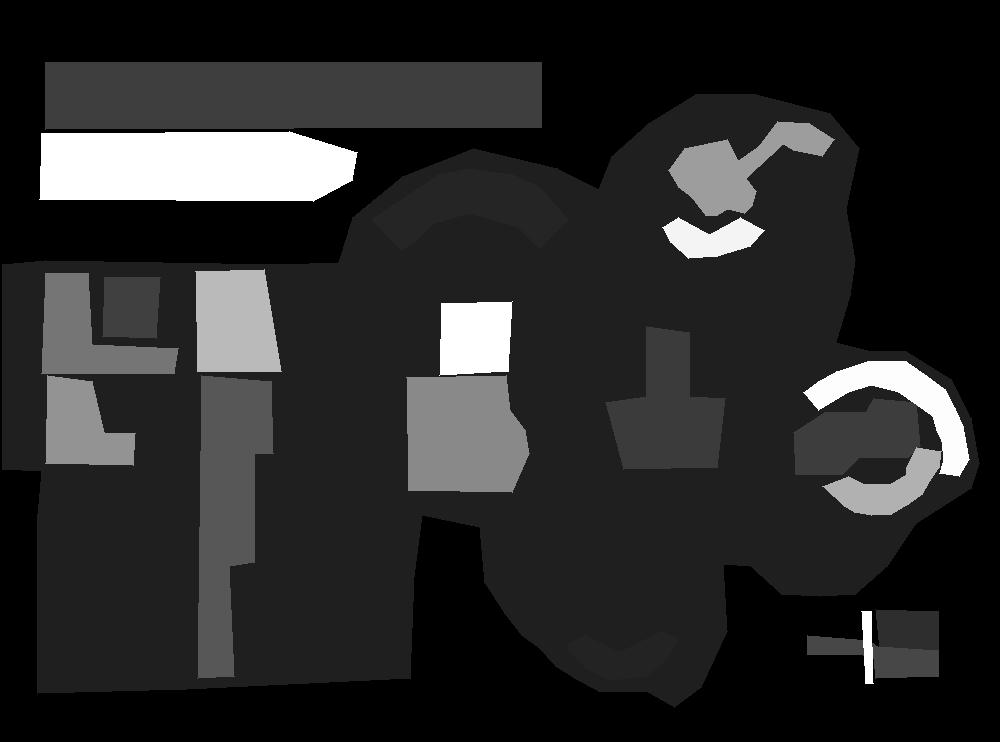}}\\
    \subfloat[Prediction of 0.5\,s]{\includegraphics[width=0.33\linewidth]{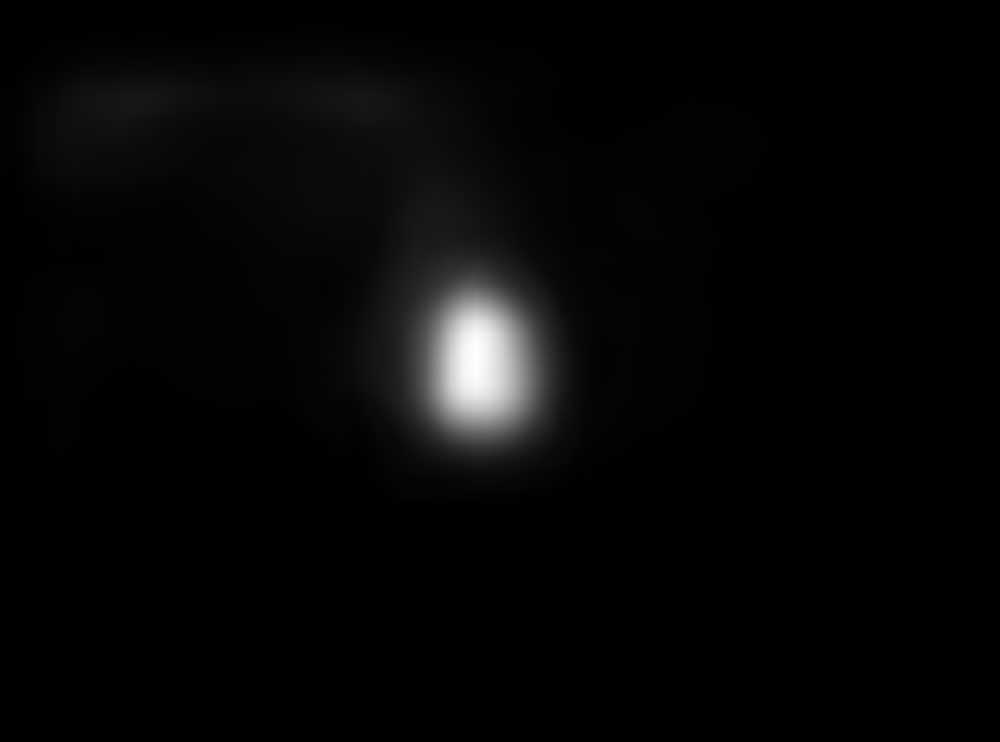}}
    \subfloat[Prediction of 3\,s]{\includegraphics[width=0.33\linewidth]{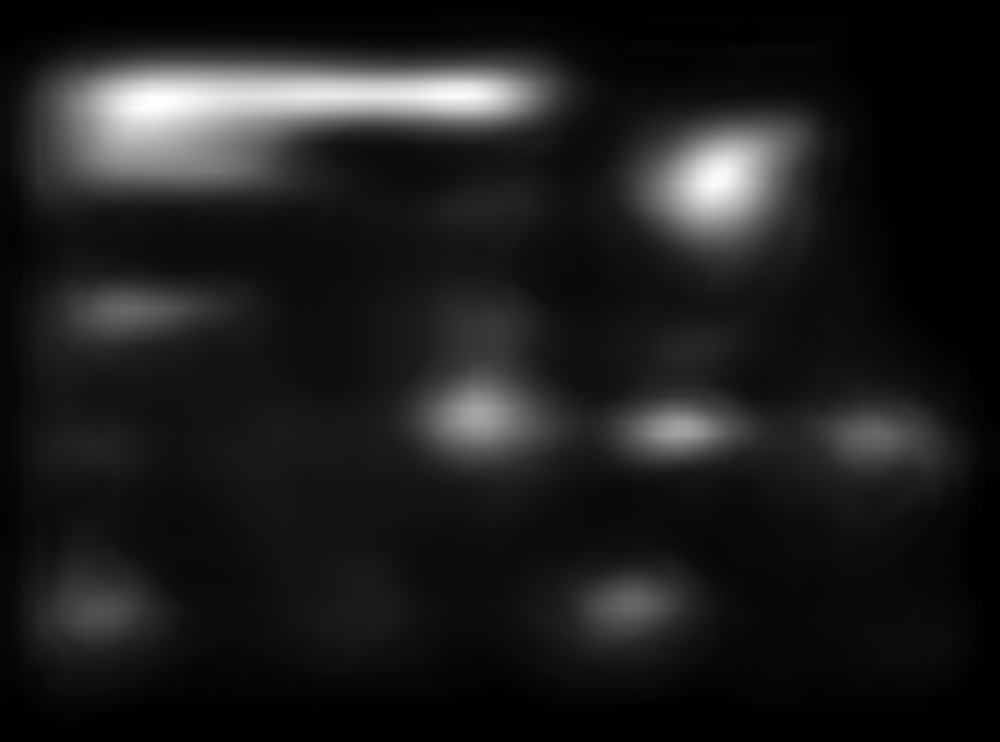}}
    \subfloat[Prediction of 5\,s]{\includegraphics[width=0.33\linewidth]{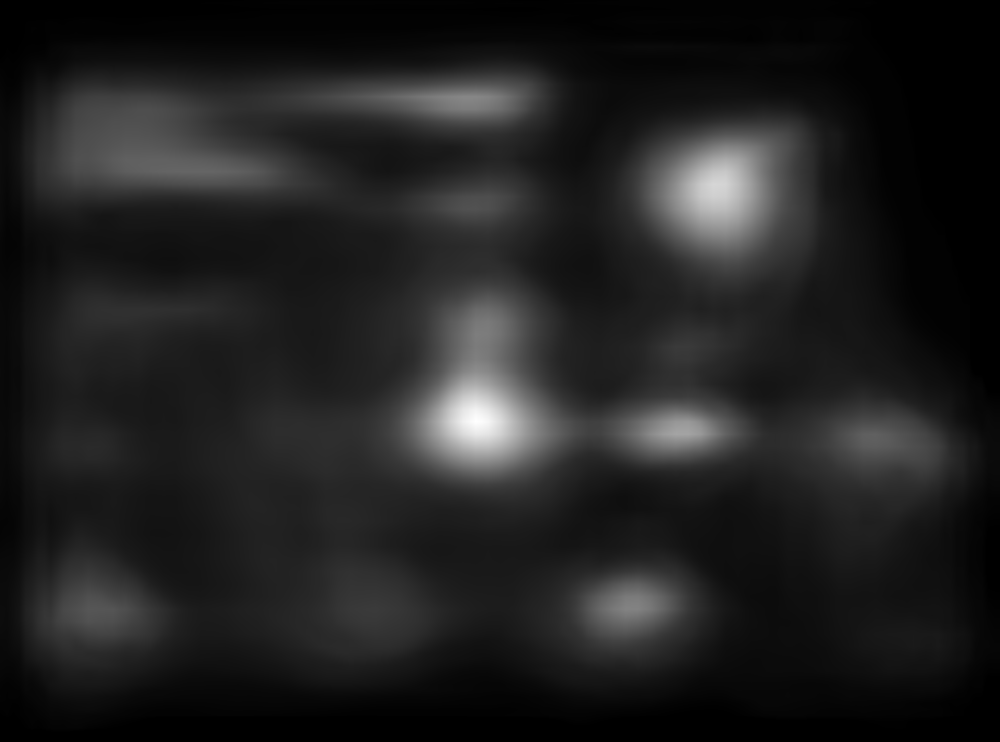}}
  \caption{One example stimulus in MASSVIS\,(a), and the corresponding saliency map of 0.5\,s\,(b), element fixation density~(EFD) maps of 3\,s\,(c), and 5\,s\,(d) time duration, and predictions of \methodSaliencyNameShort of 0.5\,s\,(e), 3\,s\,(f), and 5\,s\,(g) time duration. \methodSaliencyNameShort is able to preserve element-level information. The attention shift from Title to Data is clearly shown between (f) and (g).}
  \label{fig:example_efd}
\end{figure*}

\begin{figure*}[t]
    \centering
     \includegraphics[width=\linewidth]{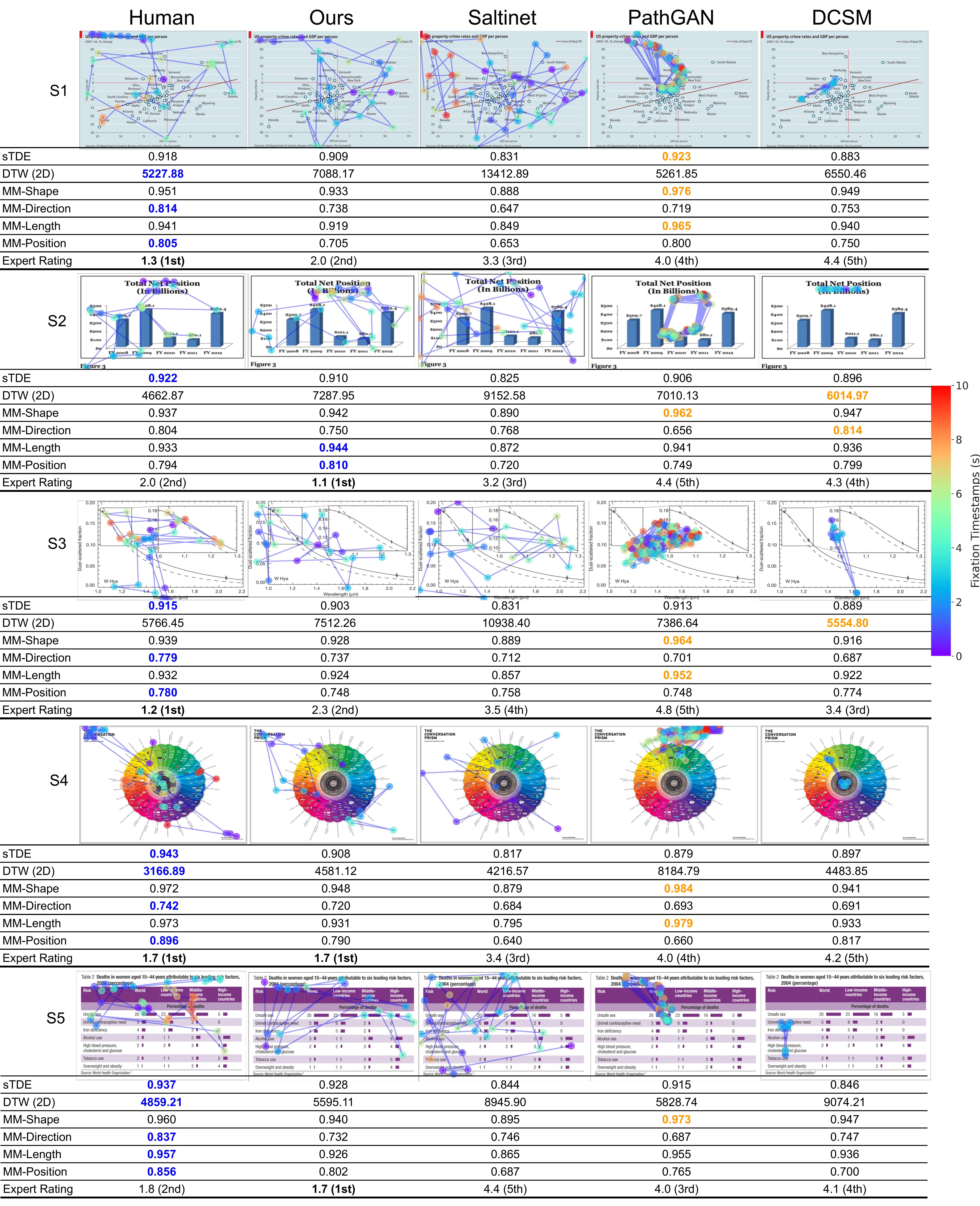}
    \caption{
    Full table of examples of mismatches between scanpath prediction performance as seen through the evaluation metrics and visualisation expert ratings.
    Each row (one visualisation from MASSVIS) shows several metrics that are contradictory to expert rating~(orange), or consistent with expert rating~(blue).}
  \label{fig:fullmetrics}
\end{figure*}

\begin{figure*}[ht]
  \centering
    \includegraphics[width=\linewidth]{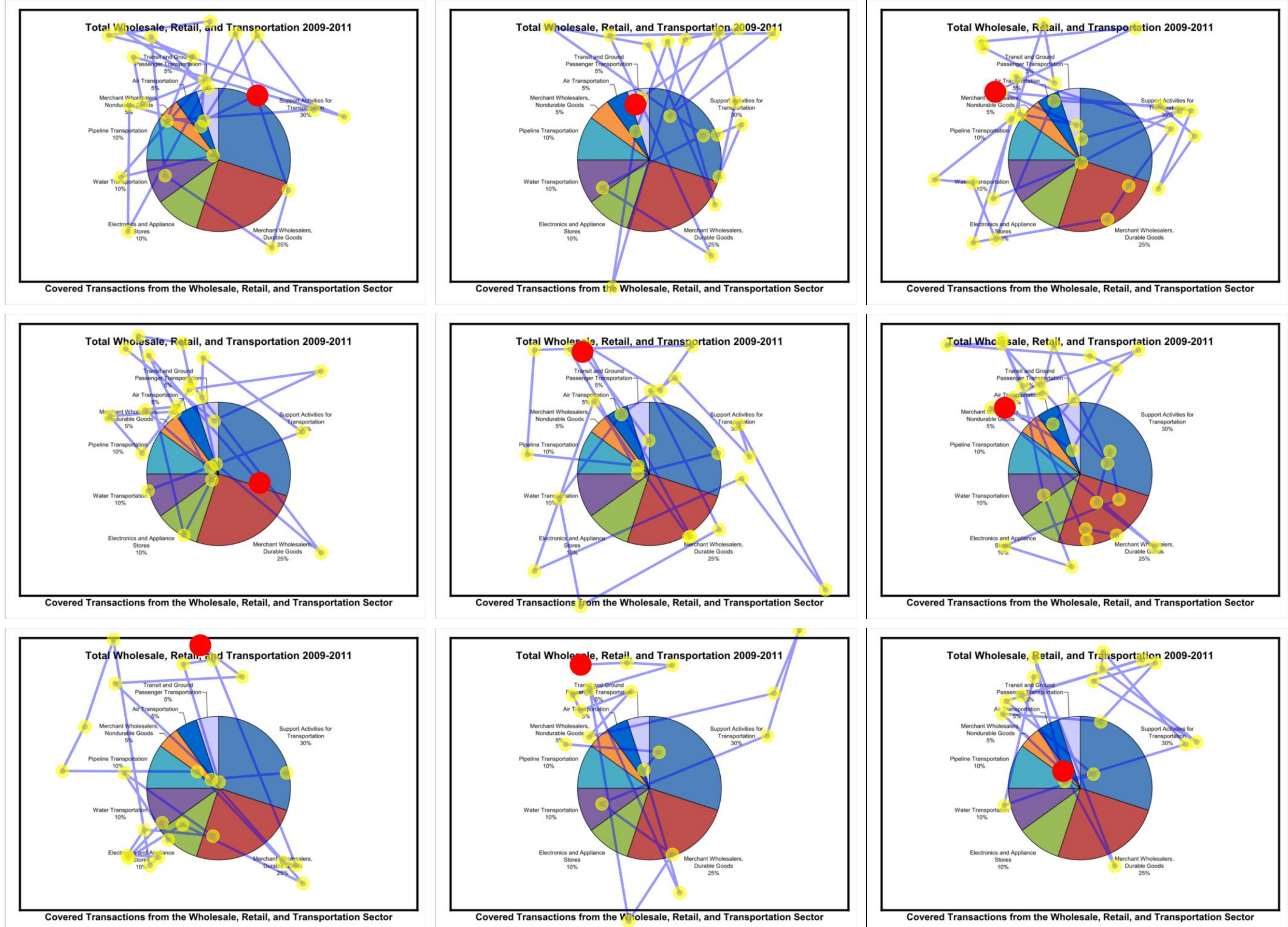}
  \caption{Scanpath predictions using UMSS~(ours) on a sample visualisation from the MASSVIS dataset.}
  \label{fig:example_1}
\end{figure*}

This document contains the implementation details of finetuning MD-SEM and training PathGAN (Section 1), distribution of scanpath length and fixation duration from the MASSVIS dataset (Figure~\ref{fig:LenDurDistribution}), The accumulated fixation distribution from the MASSVIS dataset (Figure~\ref{fig:analysis_centerbias}), transition matrices of two viewers in the MASSVIS (Figure~\ref{fig:transition_matrices}), example annotations from the MASSVIS (Figure~\ref{fig:example_annotations}), full scanpath metrics of Figure 6 in the main manuscript  (Figure~\ref{fig:fullmetrics}), example element fixation density (EFD) maps and predictions of MD-EAM in MASSVIS (Figure~\ref{fig:example_efd}), three example scanpath predictions of our UMSS model (Figure~\ref{fig:example_1}\,--\,\ref{fig:example_3}), an example questionnaire interface from our user study (Figure~\ref{fig:example_userstudy}), quantitative results on scanpath prediction for the full 10-second ground truth (Table 1), and MASSVIS~\cite{borkin2015beyond} dataset split (Table 2).

\section{Implementation Details}

\subsection{Fine-tuning MD-SEM}
We followed original setting for fine-tuning MD-SEM~\cite{fosco2020much}. 
The loss weights combination was CCM\,=\,3, KL\,=\,10, CC\,=\,-5 and NSS\,=\,-1.
Normalized Scanpath Saliency~(NSS)~\cite{peters2005components} calculates the performance of a saliency map model is defined to be the average saliency value of fixated pixels in the normalized saliency maps.
CCM is the Pearson’s Correlation Coefficient (CC)~\cite{jost2005assessing} on pairs of saliency maps at adjacent durations, which is computed as the difference between the ground truth and predicted scores~\cite{fosco2020much}.
Kullback-Leibler divergence~(KL) computes the Kullback-Leibler divergence between the empirical saliency maps and the model saliency maps after converting both of them into probability distributions\cite{kummerer2018saliency}.
Hyperparameters were batch size\,=\,8, and initial learning rate\,=\,$1E-4$.
Adam optimiser~\cite{kingma2014adam} was used for gradient descent.

\subsection{Training PathGAN}

The Root Mean Squared Propagation~(RMSprop) optimizer and Binary Cross Entropy loss with learning rate\,=\,$1E-4$, and rho\,=\,0.9, epsilon\,=\,$1E-08$, decay\,=\,$1E-07$ are used for both training and fine-tuning. During fine-tuning, we randomly mixed 5\,\% of training data from SALICON~\cite{jiang2015salicon} in each epoch to prevent forgetting~\cite{fosco2020predicting}. We trained PathGAN for 125 epochs on SALICON and 40 epochs on MASSVIS.

\begin{table}[t]
    \caption{Quantitative evaluation on MASSVIS for the full 10-second ground truth in terms of Dynamic Time Warping (DTW) and scaled Time Dimension Embedding (sTDE) metrics. Best results are shown in \textbf{bold}, second best are \underline{underlined}.}
    \label{table:scanpath_10s}
    \centering
    \begin{tabular}{ccc}
    \toprule
    Methods & DTW (2D)$~\downarrow$ & sTDE~$\uparrow$ \\
    \midrule
    Human & 8978.57 & 0.932 \\ %
    \midrule
    PathGAN~\cite{assens2018pathgan} & 10394.58 & 0.866 \\
    PathGAN-official~\cite{assens2018pathgan} & 18396.09 & 0.764\\
    DCSM~\cite{bao2020human} & \textbf{9822.26} & 0.876 \\
    Saltinet~\cite{assens2017saltinet} & 13916.36 & 0.878  \\
    DVS+Saltinet~\cite{matzen2017data, assens2017saltinet} & 13556.94 & 0.884 \\
    MDSEM+Saltinet~\cite{fosco2020much, assens2017saltinet} & 13763.52 & \underline{0.889}\\
    UMSS~(Ours) & \underline{10040.11} & \textbf{0.903} \\
    \bottomrule
    \end{tabular}

\end{table}

\begin{table}[t]
    \caption{MASSVIS~\cite{borkin2015beyond} Dataset split by visualisation source and type.}
    \label{table:massvis_split}
    \centering
    \begin{tabular}{cccc}
    \toprule
    & & Train & Evaluation \\
    \midrule
    \multirow{4}{*}{Source} & Government & 83~(25.4\%) & 17~(25.8\%) \\
    & Infographics & 77~(23.5\%) & 15~(22.7\%) \\
    & News & 101~(30.9\%) & 21~(31.8\%) \\
    & Scientific & 66~(20.2\%) & 13~(19.7\%) \\
    \midrule
    \multirow{6}{*}{Type} & Bar & 67~(20.5\%) & 17~(25.8\%) \\
    & Pie & 11~(3.4\%) & 5~(7.6\%) \\
    & Line & 57~(17.4\%) & 6~(9.1\%) \\
    & Scatter & 13~(4.0\%) & 2~(3.0\%) \\
    & Table & 28~(8.6\%) & 4~(6.1\%) \\
    & Combination & 18~(5.5\%) & 4~(6.1\%) \\
    & Other & 133~(40.7\%) & 28~(42.4\%) \\
    \midrule
    & Sum & 327~(100\%) & 66~(100\%) \\
    \bottomrule
    \end{tabular}
\end{table}

\begin{figure*}[ht]
  \centering
    \includegraphics[width=\linewidth]{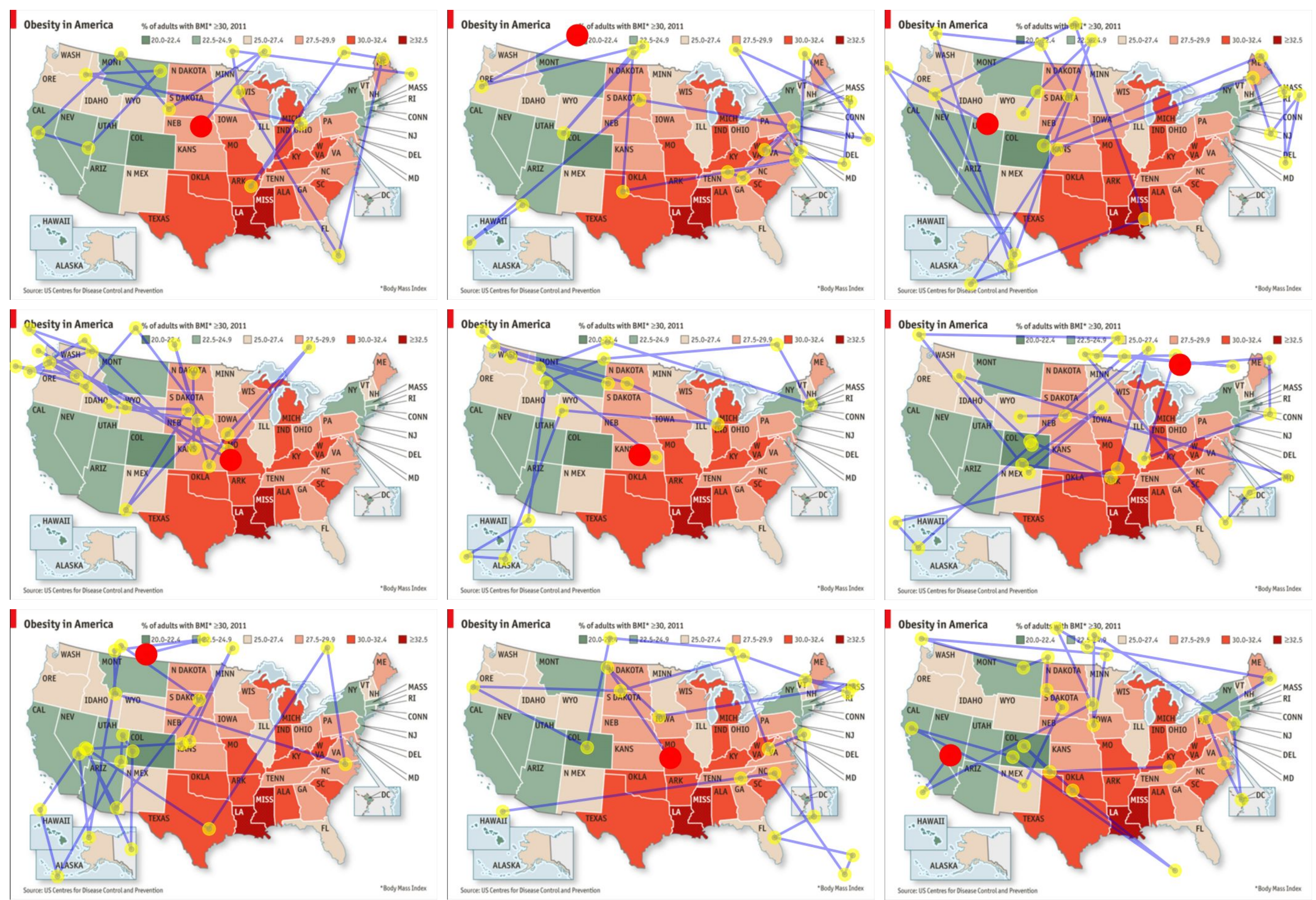}
  \caption{Scanpath predictions using UMSS~(ours) on a sample visualisation from the MASSVIS dataset.}  \label{fig:example_2}
\end{figure*}

\begin{figure*}[ht]
  \centering
    \includegraphics[width=\linewidth]{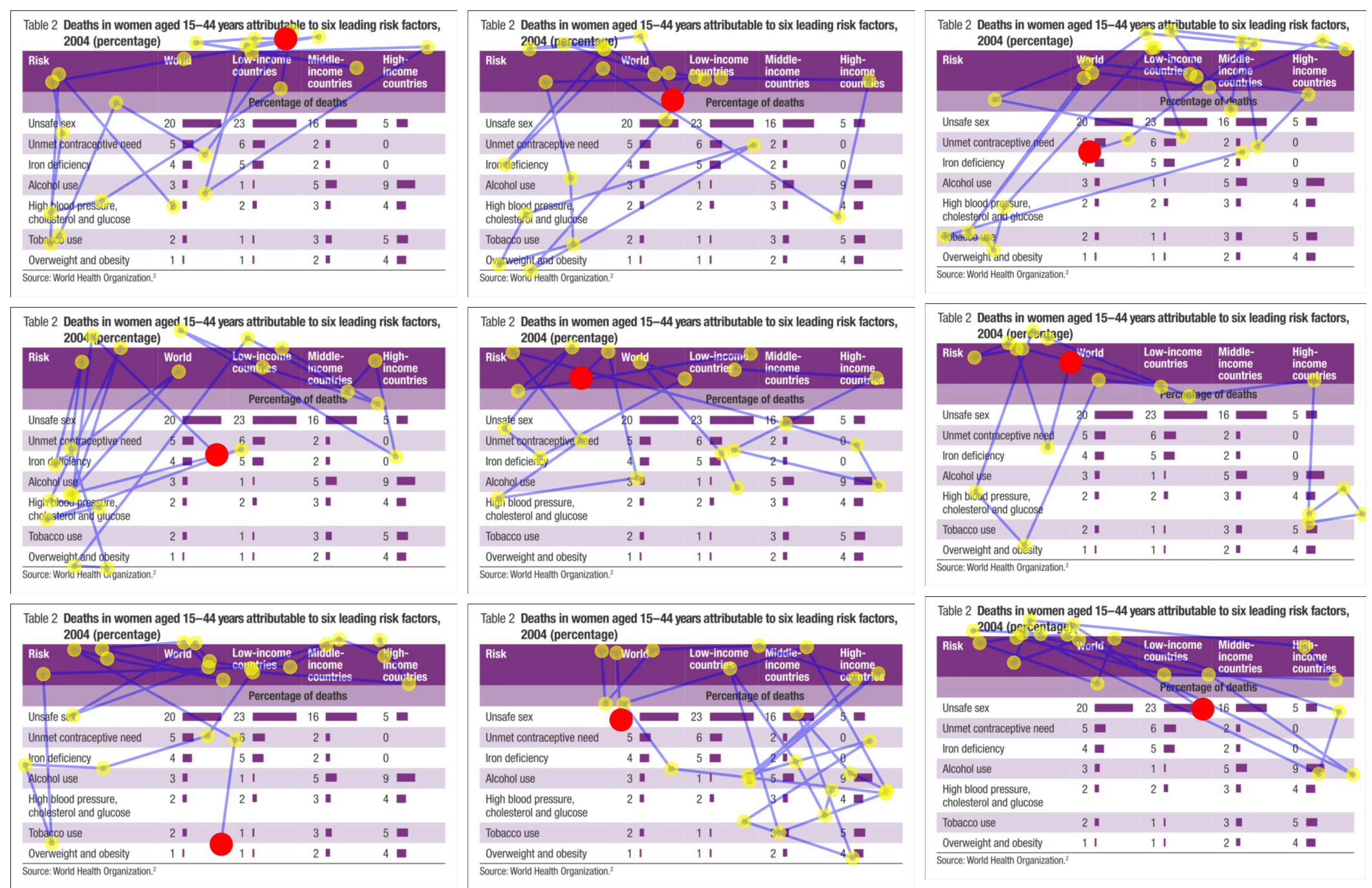}
  \caption{Scanpath predictions using UMSS~(ours) on a sample visualisation from the MASSVIS dataset.}  \label{fig:example_3}
\end{figure*}

\begin{figure*}[ht]
  \centering
    \includegraphics[width=0.9\linewidth]{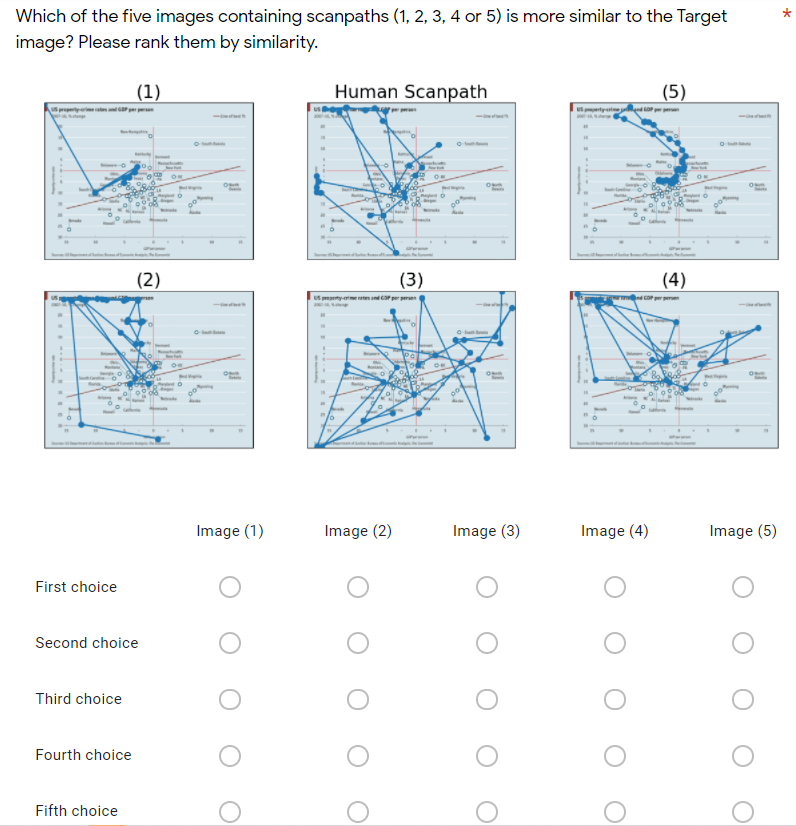}
  \caption{An example questionnaire interface of one trial (out of 40) from our user study with visualisation experts. Scanpaths were shown to the study participants as GIFs. Fixations and saccades were drawn sequentially on the image. At the end of one loop, the visualisation paused for a short period of time until a new loop started to allow subjects to compare all the scanpaths. Study participants had to rank the five options in order of their similarity when compared to one ground-truth, human scanpath. The presentation order of the baselines (1, 2, 3, 4, and 5) was counterbalanced according to a latin-square study design.}
  \label{fig:example_userstudy}
\end{figure*}

\clearpage

\bibliographystyle{IEEEtranN}
\bibliography{supplementary}